%% file: acl_latex.tex
\documentclass[11pt]{article}

\usepackage[preprint]{acl}

\usepackage{times}
\usepackage{latexsym}

\usepackage[T1]{fontenc}

\usepackage[utf8]{inputenc}

\usepackage{microtype}

\usepackage{inconsolata}

\usepackage{graphicx}

\usepackage{booktabs} 
\usepackage{footnotehyper}
\usepackage{hyperref}
\usepackage{savefnmark}
\usepackage{tablefootnote}

\usepackage{algorithm}
\usepackage{algorithmic}
\usepackage{amsmath}
\usepackage{amssymb}
\usepackage{mathtools}
\usepackage{amsthm}
\usepackage{subcaption}
\usepackage{utfsym}
\usepackage{tcolorbox}
\usepackage{pifont}
\usepackage{makecell} 

\tcbuselibrary{breakable}  
\tcbuselibrary{skins}      
\definecolor{PineGreen}{HTML}{007C4A} 
\definecolor{OrangeRed}{HTML}{FF4500} 
\definecolor{mycolor}{gray}{0.92}    

\usepackage[capitalize,noabbrev]{cleveref}

\theoremstyle{plain}

\theoremstyle{definition}

\theoremstyle{remark}

\usepackage{dcolumn, array, booktabs}
\usepackage{multirow}
\usepackage{cuted}
\usepackage[table]{xcolor} 
\usepackage[dvipsnames]{xcolor} 
\definecolor{skyblue}{RGB}{210,235,255}
\definecolor{lightcoral}{RGB}{255,228,225}
\definecolor{lemonchiffon}{RGB}{255,250,205}
\definecolor{labelRed}{RGB}{255, 0, 0}              
\definecolor{agentBlue}{RGB}{0, 112, 192}

\newtcolorbox{definitionbox}{
  colback=gray!10,
  colframe=gray!80,
  boxrule=1pt,
  width=\textwidth,
  enlarge left by=0mm,
  enlarge right by=0mm,
  float*=t,  
  boxsep=1pt,
  arc=2.5mm,
  halign=left,
}

\title{Paper2Rebuttal: A Multi-Agent Framework for Transparent Author Response Assistance}

\author{
    {\bf Qianli Ma}\footnotemark[1] \quad
    {\bf Chang Guo}\footnotemark[1] \quad
    {\bf Zhiheng Tian}\footnotemark[1] \quad 
    {\bf Siyu Wang}\quad 
    {\bf Jipeng Xiao}\quad \\ 
    {\bf Yuanhao Yue} \quad
    {\bf Zhipeng Zhang}\footnotemark[2] \\
    AutoLab, School of Artificial Intelligence, Shanghai Jiao Tong University \\
    \normalsize
    \texttt{\{mqlqianli,zhipengzhang\}@sjtu.edu.cn} \\ \\
    Project Page: \href{https://mqleet.github.io/Paper2Rebuttal_ProjectPage/}
    {\textcolor{PineGreen}{https://Paper2Rebuttal.github.io}} \\
    HF Demo: \href{https://huggingface.co/spaces/Mqleet/RebuttalAgent}
    {\textcolor{PineGreen}{https://huggingface.co/spaces/RebuttalAgent}} \\
}

\usepackage{tcolorbox} 
\tcbuselibrary{breakable}
\usepackage{listings}  
\usepackage{xcolor}    
\usepackage{enumitem}  

\begin{document}
\maketitle

{
\renewcommand{\thefootnote}{\fnsymbol{footnote}}
\footnotetext[1]{Equal Contribution.} \footnotetext[2]{Corresponding Author.}
}

\input{sec/0_abstract}
\input{sec/1_introduction_1}
\input{sec/2_related}
\input{sec/3_system_1}

\input{sec/4_benchmark}
\input{sec/5_exp}
\input{sec/6_conclusion}

\bibliography{custom}

\appendix
\input{sec/X_appendix}

\end{document}

%% file: sec/0_abstract.tex
\begin{abstract}

Writing effective rebuttals is a high-stakes task that demands more than linguistic fluency, as it requires precise alignment between reviewer intent and manuscript details. Current solutions typically treat this as a direct-to-text generation problem, suffering from hallucination, overlooked critiques, and a lack of verifiable grounding. To address these limitations, we introduce \textsc{RebuttalAgent}, the first multi-agents framework that reframes rebuttal generation as an evidence-centric planning task. Our system decomposes complex feedback into atomic concerns and dynamically constructs hybrid contexts by synthesizing compressed summaries with high-fidelity text while integrating an autonomous and on-demand external search module to resolve concerns requiring outside literature. By generating an inspectable response plan before drafting, \textsc{RebuttalAgent} ensures that every argument is explicitly anchored in internal or external evidence. We validate our approach on the proposed \textsc{RebuttalBench} and demonstrate that our pipeline outperforms strong baselines in coverage, faithfulness, and strategic coherence, offering a transparent and controllable assistant for the peer review process. 

\end{abstract}

%% file: sec/1_introduction_1.tex
\section{Introduction}
\label{sec:intro}

The rebuttal phase represents a decisive juncture in the peer review lifecycle where authors must address critiques through evidence-backed clarifications and actionable manuscript revisions. This undertaking extends far beyond simple textual composition. It requires a rigorous synthesis process in which authors must accurately decipher reviewer intent while ensuring every response is firmly anchored in verifiable manuscript details. The inherent difficulty of this multi-step reasoning is amplified by the strict turnaround windows typical of top-tier venues. Authors are frequently forced to reconcile the need for meticulous verification with urgent deadlines, leaving little room for hallucination or ambiguity. 


\input{fig/fig1-intro}

In response to these intense cognitive and temporal demands, Large Language Models (LLMs) have emerged as promising assistants for scientific writing~\cite{wang2024autosurvey} and peer-review communication~\cite{gao2024reviewer2,zhu2025deepreview,luagent}. Current approaches generally fall into two paradigms. The \textit{direct-to-text} generation paradigm typically involves models that are supervised fine-tuned (SFT) on paper-response pairs (Fig.~\ref{fig:overview}\textcolor{darkblue}{a}). While straightforward, this approach is fundamentally flawed because it trains models to memorize specific, non-transferable experimental outcomes rather than the underlying logic of formulating a strategic response. Consequently, these models are prone to hallucination, frequently fabricating experimental results or over-commit to unverified claims instead of reasoning about the actual content of the manuscript. The second paradigm relies on interactive sessions with proprietary chat-LLMs such as GPT or Gemini (Fig.~\ref{fig:overview}\textcolor{darkblue}{b}). While these high-capability models can offer superior reasoning, the workflow is notoriously inefficient and opaque. Authors are forced to engage in lengthy, multi-turn prompting to guide the model, which consumes valuable time that could be spent on verification. Furthermore, critical intermediate steps like concern parsing and evidence retrieval remain concealed behind the chat interface. This lack of transparency makes the process difficult to audit and renders the output quality heavily dependent on the prompting expertise of the user.

In this paper, we reframe rebuttal assistance as a \textit{decision and evidence organization problem} with explicit constraints, rather than the free-form text generation tasks. Specifically, a reliable system must satisfy four critical requirements: \textbf{(i) Comprehensive Coverage,} tracking every reviewer's concern without omission; \textbf{(ii) Strict Faithfulness,} adhering to the submitted manuscript without hallucinating technical details; \textbf{(iii) Verifiable Grounding,} linking major statements to specific internal passages or external references; and \textbf{(iv) Global Consistency,} maintaining a unified stance and avoiding conflicting commitments across different responses. To instantiate this view, we propose \textsc{RebuttalAgent}, a multi-agent system that enforces a novel "verify-then-write" workflow to overcome the opacity of previous two paradigms, shown in Fig.~\ref{fig:overview}\textcolor{darkblue}{c}.

Instead of rushing to generation, our architecture explicitly decouples reasoning from drafting by producing verifiable intermediate artifacts. The process begins by atomizing unstructured reviews into discrete concerns to guarantee comprehensive coverage, followed by a dual-source evidence construction phase that synthesizes high-fidelity manuscript passages and citation-ready external briefs to strictly ground every claim. Crucially, we introduce a strategic planning stage that audits the response logic for global consistency and commitment safety before any text is drafted, ensuring that concessions made to one reviewer do not contradict the overall stance. By exposing these structured artifacts through human-in-the-loop checkpoints, \textsc{RebuttalAgent} transforms rebuttal writing from a black-box generation task into a transparent, author-controlled collaboration.

We evaluate \textsc{RebuttalAgent} through an author-centric lens, prioritizing practical usability and reliability over mere text fluency. Specifically, we assess performance across four rigorous dimensions: \textbf{coverage} of reviewer concerns, \textbf{traceability} of evidence sources, \textbf{global coherence} of the argumentative stance, and overall \textbf{argumentation quality}. Experimental results on our proposed benchmark demonstrate that our pipeline consistently outperforms previous "direct-to-text" baselines and chat-LLMs on these critical metrics. By delivering structured, verifiable assistance, \textsc{RebuttalAgent} significantly reduces the cognitive burden of rebuttal writing while ensuring authors remain the ultimate arbiters of their scientific defense.

Our contributions are: \ding{171} We formulate rebuttal assistance as a decision-and-evidence organization problem and propose \texttt{RebuttalAent}, a multi-agent system with explicit verification and human-in-the-loop checkpoints. \ding{171} We introduce concern-conditioned context construction and on-demand evidence synthesis to produce point-specific, verifiable support under realistic context limits. \ding{171} We construct a benchmark and establish an author-centric evaluation protocol, demonstrating that our approach outperforms baselines in coverage, traceability, and coherence.

%% file: fig/fig1-intro.tex
\begin{figure}[!t]
            \centering
            \includegraphics[width=\linewidth]{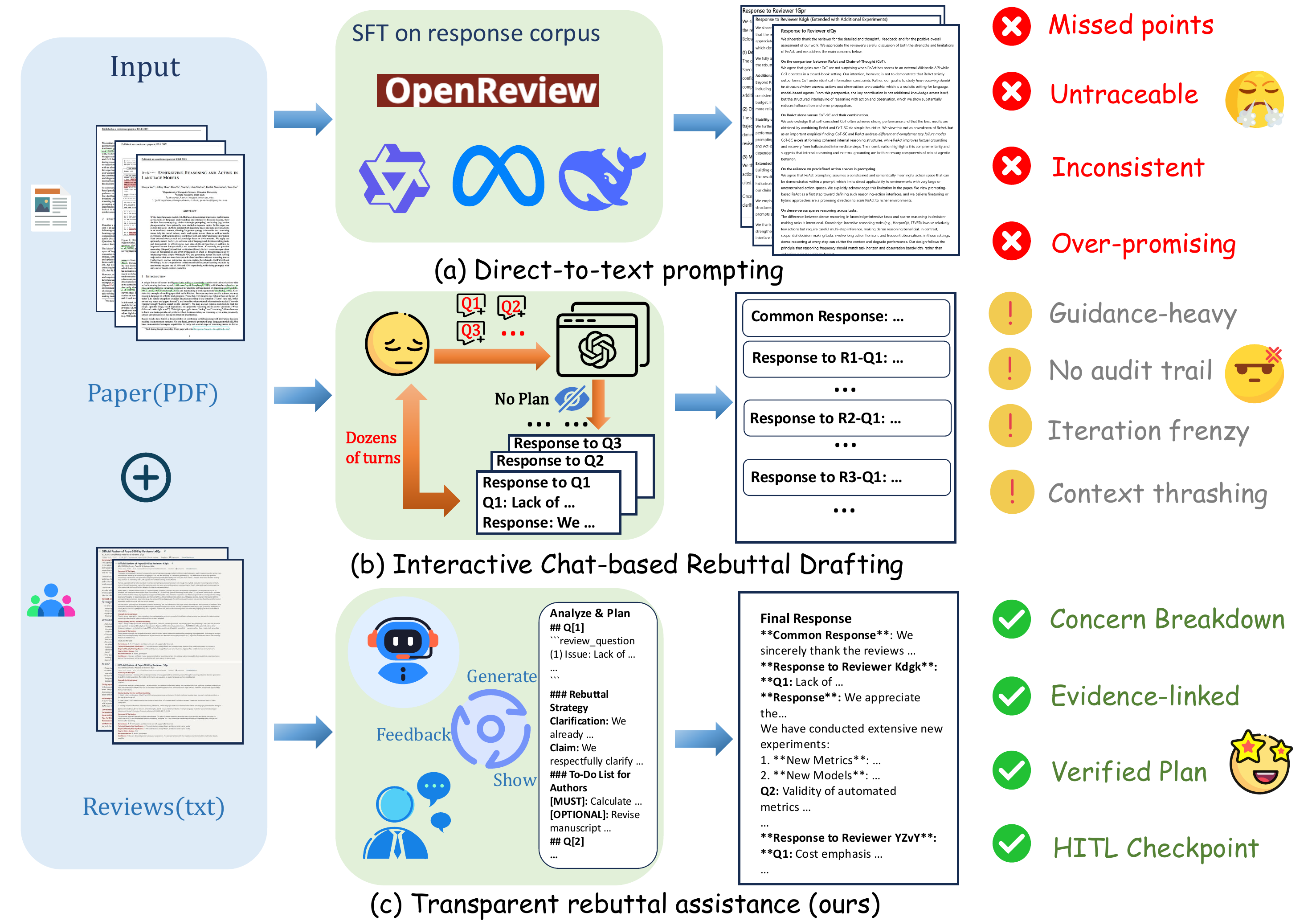}
            \caption{\textbf{Overview of our work.} Given a manuscript and reviews, \textbf{(a)} direct text generation (SFT on peer-review corpora) often fabricates experiment results and prone to hallucination. \textbf{(b)} Interactive prompting with chat-LLMs depends on manual concern feeding and many iterations. \textbf{(c)} RebuttalAgent reframes rebuttal writing as a decision-and-evidence organization problem, performing concern breakdown, query-conditioned internal and external evidence construction, and strategy-level plan verification with human-in-the-loop checkpoints before drafting the final response.}
            \vspace{-15pt}
     \label{fig:overview}
\end{figure}

%% file: sec/2_related.tex
\section{Related Works}
\label{sec:related}
\input{fig/fig2-method}

\noindent\textbf{LLM Agents.}~
LLMs~\cite{openai2025gpt5,team2023gemini} were initially valued for fluent generation, but real deployments revealed a mismatch between writing well and completing complex tasks reliably. When goals require multi-step planning, fresh evidence, and interaction with external systems, purely parametric generation can accumulate errors and hallucinations, motivating a shift toward intelligent “agents” embedded in dynamic, goal-directed frameworks that  plan and act with external tools and environments.
Recent work~\cite{yao2023reactsynergizingreasoningacting,yao2023treethoughtsdeliberateproblem} shows that combining reasoning traces with concrete actions (\textit{e.g.}, search tool) improves robustness in long-horizon tasks and reduces hallucinations.
Modern agents often incorporate deliberation and search~\cite{wei2023chainofthoughtpromptingelicitsreasoning,wang2024autosurvey}, learned tool-use policies~\cite{schick2023toolformerlanguagemodelsteach,patil2023gorillalargelanguagemodel}, and memory or reflection from execution feedback~\cite{shinn2023reflexionlanguageagentsverbal,zhang2024surveymemorymechanismlarge}.
Multi-agent frameworks further enable role specialization and structured collaboration~\cite{Wang_2024,wu2023autogenenablingnextgenllm,autopage,luagent,d2024marg}, while benchmarks such as AgentBench~\cite{liu2025agentbenchevaluatingllmsagents}, WebArena~\cite{zhou2024webarenarealisticwebenvironment}, and GAIA~\cite{mialon2023gaiabenchmarkgeneralai} evaluate real-world tool use and end-to-end task success. These advances motivate extending agentic systems from \emph{conducting} research to \emph{communicating} it, \textit{e.g.}, retrieving evidence, organizing words and iteratively refining rebuttals.


\noindent\textbf{AI Assisted Peer Review.}~
Peer review stands as the cornerstone of research quality yet faces significant strain from the exponential growth in conference submissions. This pressure has catalyzed the adoption of LLMs to maintain efficiency and decision reliability across the review pipeline~\cite{gao2024reviewer2,luagent,zhu2025deepreview,zhang2025re}. Within this process, the author rebuttal phase holds unique value for rectifying misunderstandings and influencing borderline decisions~\cite{gao2019does}. To operationalize this complex interaction, researchers have developed datasets like DISAPERE~\cite{kennard2021disapere} and APE~\cite{cheng2020ape} for argument alignment alongside comprehensive corpora like $Re^2$~\cite{zhang2025re}. While recent efforts employ argumentative strategies~\cite{purkayastha2023exploring} or multi-agent simulations~\cite{yu2025researchtownsimulatorhumanresearch,jin2024agentreview} to model this workflow, they predominantly treat rebuttal generation as a single-step prompt-to-text task. As illustrated in Fig.~\ref{fig:overview}\textcolor{darkblue}{a}, these methods overlook the critical need for explicitly decomposing concerns and planning evidence-based responses.

%% file: fig/fig2-method.tex
\begin{figure*}[!t]
            \centering
            \includegraphics[width=\textwidth]{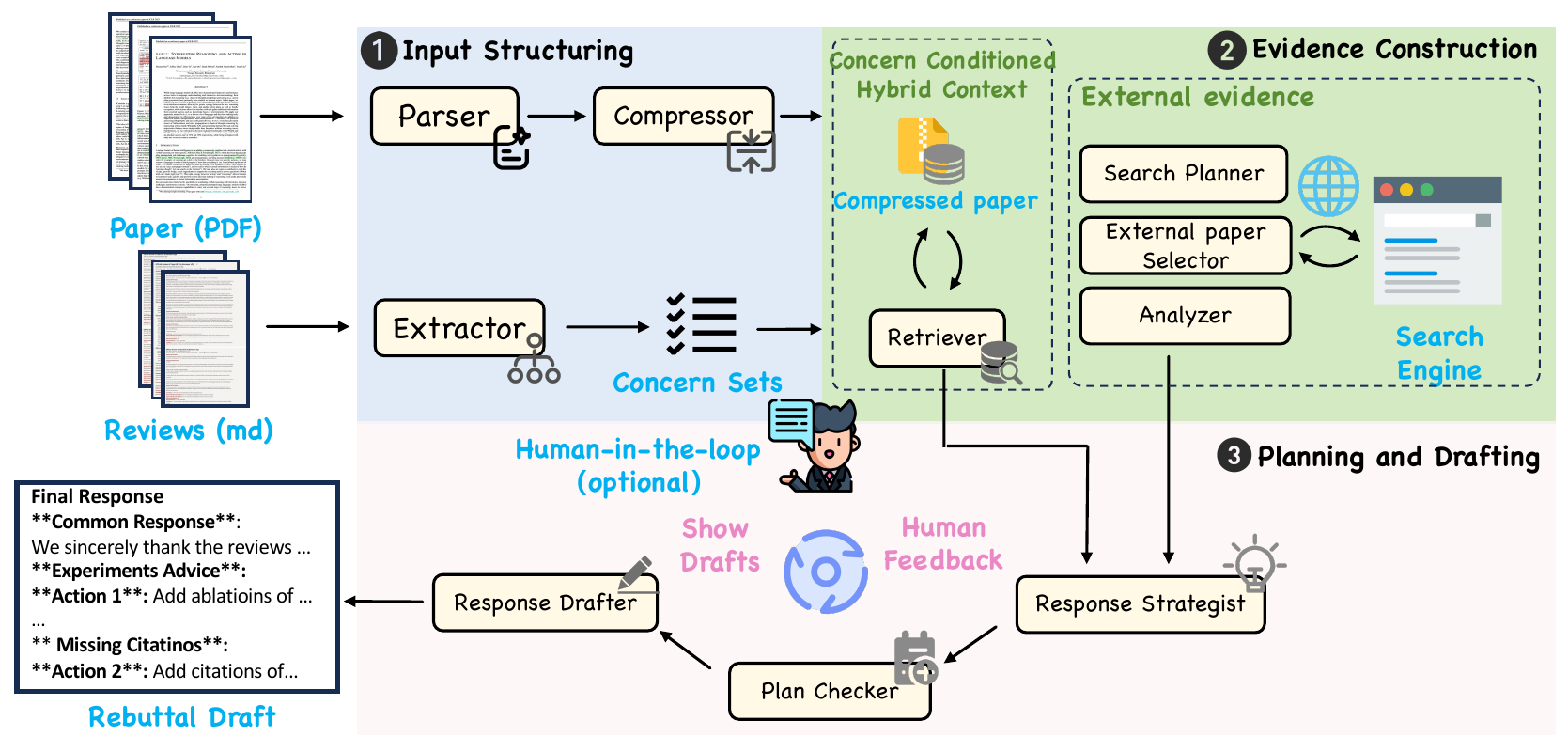}
            \caption{\textbf{Overview of RebuttalAgent.} Given a manuscript (PDF) and reviewer comments, the system \textbf{(1)} structures inputs by parsing and compressing the paper with fidelity checks and extracting atomic reviewer concerns with coverage checks; \textbf{(2)} builds concern-conditioned evidence by constructing a query-specific hybrid manuscript context and, when needed, retrieving and summarizing external literature into citation-ready briefs; and \textbf{(3)} generates an inspectable, evidence-linked response plan that is checked for consistency and commitment safety, incorporates optional author feedback, and is then realized into a formal rebuttal draft.}
            \vspace{-10pt}
     \label{fig:pipeline}
\end{figure*}

%% file: sec/3_system_1.tex
\section{RebuttalAgent}
\label{sec:rebutsystem}

\textsc{RebuttalAgent} operates as a multi-agent framework that transforms the rebuttal process into a structured and inspectable workflow. By generating evidence-linked intermediate artifacts before drafting the final text, the system ensures that the output remains grounded and controllable. Fig.~\ref{fig:pipeline} illustrates how the architecture decomposes complex reasoning into specialized agents paired with lightweight checkers. This design exposes critical decision points and allows authors to retain full responsibility for the strategic stance and final wording. The pipeline initiates by distilling the manuscript into a structured summary and extracting atomic reviewer concerns to enable stable long-context reasoning (Sec.~\ref{subsec:input_parse}). Guided by these atomic concerns, the system constructs evidence bundles by retrieving relevant high-fidelity excerpts from the original manuscript and augmenting them with verifiable external literature via web search (Sec.~\ref{subsec:evidence}). The workflow concludes by synthesizing an explicit response plan that outlines the arguments and evidence links. Authors refine this plan through a human-in-the-loop mechanism before the system produces a formal rebuttal letter compliant with academic conventions (Sec.~\ref{subsec:drafting}).


\subsection{Manuscript and Review Structuring}
\label{subsec:input_parse}


The pipeline commences by distilling the raw manuscript and reviews into condensed representations optimized for downstream reasoning. This approach addresses the dual challenges of efficiency and controllability as effective rebuttals demand repeated access to fine-grained evidence scattered throughout the paper. Since processing the full manuscript directly is often costly and brittle due to context limitations, our compact format minimizes token overhead while improving retrieval precision. It serves as a navigational anchor that allows subsequent modules to selectively access high-fidelity excerpts from the original text only when precise evidence is necessary.


\noindent\textbf{Dense Manuscript to Compact Representation.} The transformation begins as a parser agent converts the manuscript PDF into a paragraph-indexed format to preserve structural integrity and facilitate targeted lookups. A compressor agent subsequently distills these paragraphs into a concise representation that retains essential technical statements and experimental results. This compact view functions as the primary retrieval surface and enables the system to match reviewer concerns to relevant sections with minimal token usage. To safeguard against silent information loss, a consistency checker verifies each condensed unit against its source and automatically triggers reprocessing if it detects missing claims or semantic drift.

\noindent\textbf{Complex Reviews to Actionable Atomic Concerns.} Operating in parallel with manuscript processing, an extractor agent parses raw feedback into discrete and addressable atomic concerns. This component organizes the critiques by grouping related sub-questions and assigning preliminary categories. A coverage checker subsequently validates the output for intent preservation and appropriate granularity to guarantee that substantive points remain distinct without being over-split or incorrectly merged. The resulting structured list forms the foundational unit for the subsequent evidence gathering and response planning stages.

\subsection{Evidence Construction}
\label{subsec:evidence}

With the atomic concerns established, the system generates targeted evidence bundles to ensure that every argument remains traceable to specific facts. This strategy contrasts sharply with the direct generation approaches depicted in Fig.~\ref{fig:overview}\textcolor{darkblue}{a} that bypass explicit grounding. By prioritizing evidence construction over immediate text generation, our pipeline anchors each concern in verifiable sources and ensures that the downstream planning and drafting stages operate on validated information.

\noindent\textbf{Atomic Concern Conditioned Hybrid Context.} The system identifies the most pertinent sections by searching within the compressed manuscript representation (Sec.3.1) for each atomic concern. It then selectively expands these focal points by retrieving the corresponding raw text to replace the specific condensed units while retaining the rest of the document in its summarized form. This approach yields an atomic concern conditioned hybrid context that integrates the efficiency of the compressed view with the precision of the original text. Such a structure enables the system to support its reasoning with exact quotations and detailed evidence without overwhelming the context window.

\noindent\textbf{On-Demand External Evidence.} While the hybrid context effectively grounds responses in the authors' own work, certain reviewer critiques necessitate evidence beyond the manuscript boundaries. To address scenarios such as novelty disputes or requests for broader positioning where internal data is insufficient, the system augments the evidence bundle with external support. A search planner initiates this expansion by formulating a targeted search strategy, while a subsequent retrieval step gathers candidate papers via scholarly search tools\footnote{https://export.arxiv.org/api/query}. A screening agent then filters these candidates for relevance and utility to ensure high-quality input. The pipeline concludes this phase by parsing the selected works into a structured evidence brief that highlights key claims and experimental comparisons to provide citation-ready material for the subsequent planning and drafting stages.

\input{fig/fig3-bench}

\subsection{Planning and Drafting}
\label{subsec:drafting}
A critical failure of the direct-to-text pipeline is its tendency to hallucinate experimental results when addressing empirical critiques. Our system overcomes this by implementing a bifurcated reasoning strategy that strictly distinguishes between interpretative defense and necessary intervention. For concerns resolvable through existing data, the Strategist Agent synthesizes arguments directly from the hybrid context and anchors them in the manuscript text. In contrast, when the system detects a demand for new experiments or baselines, it explicitly inhibits result generation and instead produces concrete \emph{Action Items} framed as recommendations (see cases in App.~\ref{app:case_study}). This design prevents the fabrication of outcomes by forcing a structural pause where authors must verify or perform the suggested tasks. The resulting plan serves as an interactive human-in-the-loop checkpoint that allows authors to actively refine the strategic logic rather than merely accepting or rejecting proposals. Users can modify the scope of action items or correct the reasoning path to ensure the strategy aligns perfectly with their capabilities and intent. Only after the author validates these strategic decisions does the Drafter Agent convert the plan into a final response to ensure that every claim remains grounded in reality.
Optionally, the drafter can also produce a submission-style rebuttal draft from the validated plan, but it renders any yet-to-be-conducted experiments as explicit placeholders (\textit{e.g.}, \texttt{[TBD]}).
Authors can then fill in these placeholders after completing the recommended action items, keeping the draft faithful.

%% file: fig/fig3-bench.tex
\begin{figure*}[!ht]
            \centering
            \includegraphics[width=\textwidth]{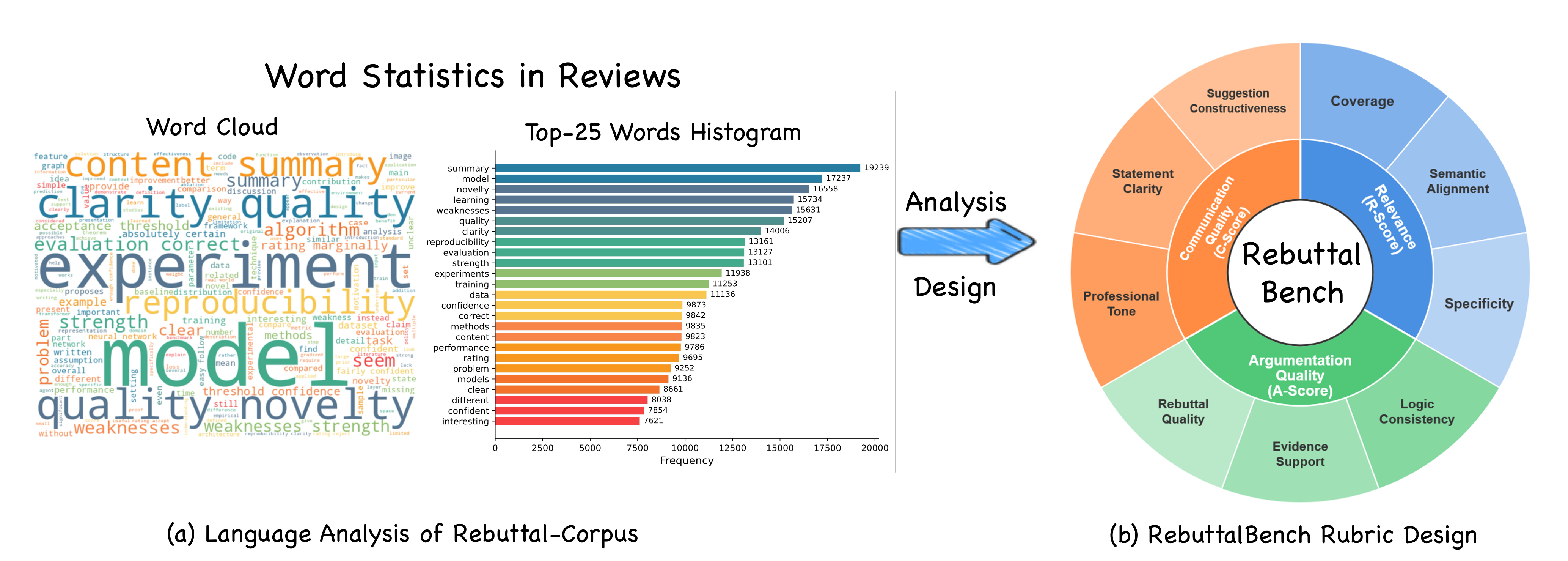}
            \vspace{-20pt}
            \caption{\textbf{RebuttalBench statistics and rubric design.} \textbf{(a)} Word-cloud and top-word histogram of reviews in \textsc{RebuttalBench-Corpus}, highlighting recurring reviewer emphases (\textit{e.g.}, clarity, novelty, reproducibility). \textbf{(b)} Motivated by these signals, \textsc{RebuttalBench} evaluates rebuttals with a rubric that mirrors these concerns, scoring \emph{Relevance}, \emph{Argumentation Quality}, and \emph{Communication Quality} rather than fluency alone.}
            \vspace{-15pt}
     \label{fig:bench}
\end{figure*}

%% file: sec/4_benchmark.tex
\section{RebuttalBench}
\label{sec:bench}



Standard evaluation metrics for text generation fail to capture the strategic nuance and factual precision required in peer review rebuttals. Therefore, we introduce \textsc{RebuttalBench} as a specialized benchmark derived from real-world OpenReview interactions. This dataset moves beyond simple text-to-text pairs by curating high-quality review-response dyads to ensure technical density and argumentative complexity. We complement the data with a multidimensional evaluation framework that prioritizes content coverage and evidence traceability over surface-level fluency. Unlike generic instruction-following benchmarks, our protocol specifically measures how well a system identifies atomic concerns and grounds its counter-arguments in verifiable facts. This allows us to quantify the gap between the hallucination-prone outputs of standard models and the structured reasoning produced by our pipeline.

\subsection{Evaluation Dataset}
\label{subsec:eval_dataset}

\noindent\textbf{Data source.}~
To evaluate rebuttal assistance with an observable post-rebuttal signal, we curated a dataset of peer-review discussion threads from the publicly available ICLR OpenReview forum. Each instance in our benchmark pairs an initial reviewer critique with the corresponding author rebuttal and crucially includes the reviewer's follow-up response. We leverage the subsequent reviewer reaction as a decisive classification signal to partition the dataset into positive and negative samples for evaluation purposes. Positive instances are identified by follow-up comments confirming that all concerns were resolved while negative samples consist of cases where the reviewer indicated that the rebuttal failed to address the core issues.



\noindent\textbf{Filtering and corpus construction.}~
Starting from the raw peer-review discussion threads, we apply automatic filtering to retain instances with sufficiently explicit follow-up signals and remove ambiguous cases to obtain a broad and reliable evaluation pool.
This yields \texttt{RebuttalBench-Corpus}, a broad pool of 9.3K review-rebuttal pairs used for analysis and evaluation setup. (see Appendix~\ref{app:evaluation_dataset}).
To form a focused and challenging benchmark for standardized comparison, we construct \textsc{RebuttalBench-Challenge} by ranking papers according to the number of instances that exhibit both positive and negative follow-up signals, and selecting the top 20 papers with over 100 reviewers.
This strategy maximizes within-paper diversity of resolved and unresolved concerns, producing a compact test suite with realistic interaction patterns. We provide details about the filtering strategy in Appendix~\ref{app:filter}.

\noindent\textbf{Data statistics.}
Fig.~\ref{fig:bench} summarizes corpus-level characteristics of \textsc{RebuttalBench-Corpus}.
Beyond basic length and interaction statistics, we visualize reviewer language with a word cloud and top-words histogram, shown in Fig.~\ref{fig:bench}\textcolor{darkblue}{a}.
Frequent terms such as \emph{clarity}, \emph{quality}, \emph{correct(ness)}, \emph{reproducibility}, \emph{novelty}, and \emph{experiments} indicate that reviewers repeatedly emphasize exposition, claim support, and scientific rigor; these axes are also explicitly reflected in standard review forms used in OpenReview venues.
Accordingly, our rubric-based evaluation is designed to align with these recurring concerns by scoring relevance/coverage to reviewer points, strength of evidence-backed argumentation, and communication quality (\textit{e.g.}, clarity and professionalism), demonstrated in Fig.~\ref{fig:bench}\textcolor{darkblue}{b}.

\subsection{Evaluation Metrics}
\label{subsec:eval_metrics}
To systematically measure rebuttal response quality beyond surface fluency, we use an LLM-as-judge~\cite{zheng2023judging,lin2023llm} rubric with a fine-grained \textbf{0-5} scale.
The evaluation framework covers three complementary dimensions: \textit{Relevance} (R-Score), \textit{Argumentation Quality} (A-Score), and \textit{Communication Quality} (C-Score).
Each dimension contains three components (9 total). 
We calculate the average component scores within each dimension and then compute the final score.
Full component rubrics and judge prompts are provided in Appendix~\ref{app:evaluation_metric}.





\noindent\textbf{R-Score}
evaluates the extent to which the response addresses reviewer concerns with point-specific precision. It rewards outputs that cover all major points without omission and demonstrate a correct interpretation of the critique while favoring concrete actions over generic assurances

\noindent\textbf{A-Score}
measures the strength of the justification behind each claim. It requires arguments to be logically consistent and supported by appropriate evidence from the manuscript or external sources. The metric prioritizes substantive rebuttals that engage with the underlying critique rather than offering superficial restatements.

\noindent\textbf{C-Score}
captures the quality of communication and professional conduct. It assesses whether the response maintains a respectful tone and presents information with a clear structure and unambiguous language. The metric ensures the text remains constructive to facilitate a productive discussion between the reviewer and the author.

In addition to scalar scores, the evaluator outputs a brief structured diagnosis (strengths, weaknesses, and suggested improvements) for qualitative analysis. Detailed scoring standards (0-5 anchors) and implementation are provided in Appendix~\ref{app:evaluation_metric}.

%% file: sec/5_exp.tex
\section{Experiments}
\label{sec:experiment}
\input{tab/main_exp}

\subsection{Experimental Setup}
We assess the efficacy of \textsc{RebuttalAgent} by comparing it with strong closed-source LLM baselines and by ablating key components of the system.
For scalable and controlled benchmarking, all experiments in the main paper run \textsc{RebuttalAgent} in a fully automated mode without human intervention.
While human-in-the-loop checkpoints can further improve reliability and author control, they are impractical for batch evaluation at scale.
Accordingly, the reported results should be viewed as a \textit{conservative lower bound} on the system’s performance under real-world usage.

\noindent\textbf{Baselines.}
We consider four SOTA LLMs as baselines: GPT-5-mini~\cite{openai2025gpt5}, Grok-4.1-fast~\cite{xGrok}, Gemini-3-Flash~\cite{team2023gemini}, and DeepSeekV3.2~\cite{liu2025deepseek}.
We also compare with general multi-agent systems~(See App.~\ref{app:more_baselines}).
For each baseline model, we evaluate a \emph{direct-to-text generation} setting where the model produces a rebuttal conditioned on the manuscript and reviewer comments.
To ensure a fair comparison, we also instantiate \textsc{RebuttalAgent} with the same model as its foundation backbone, keeping inputs and outputs identical across conditions; differences therefore reflect the contribution of our structured pipeline rather than the underlying model choice.

\noindent\textbf{Implementation Details.}
To ensure controlled and fair comparisons, we evaluate \textsc{RebuttalAgent} and each closed-source baseline under matched model backbones.
For every baseline LLM (\textit{e.g.}, GPT-5-mini~\cite{openai2025gpt5}), we instantiate \textsc{RebuttalAgent} with the same LLM as its backbone, so that both the baseline and \textsc{RebuttalAgent} consume the same manuscript and reviewer comments and produce responses in an identical point-by-point format. Differences therefore reflect the contribution of the structured workflow rather than model capacity.
All experiments in the main paper run RebuttalAgent in a fully automated mode, and we keep decoding settings consistent across conditions for each backbone.
Finally, we adopt Gemini-3-Flash~\cite{team2023gemini} as a unified LLM judge for all systems and ablations.
Full prompt templates and evaluation prompts are provided in Appendix~\ref{app:evaluation_metric} and Appendix~\ref{app:prompt-templates}.


\subsection{Main Results}
\noindent\textit{\textbf{Obs. 1:}} \textbf{RebuttalAgent consistently outperforms strong closed-source LLMs.}
As shown in Tab.~\ref{tab:main_results}, under fair comparisons where \textsc{RebuttalAgent} and LLM baselines share the same base models, \textsc{RebuttalAgent} achieves consistent improvements across all evaluation dimensions on \textsc{RebuttalBench}.
The largest gains are observed in \textit{Relevance} and \textit{Argumentation Quality}.
Across matched base models, \textsc{RebuttalAgent} improves \textit{coverage} by up to \textbf{+0.78} for DeepSeekV3.2 and \textit{specificity} by up to \textbf{+1.33} for GPT5-mini, and strengthens argumentation with up to \textbf{+0.63} higher \textit{rebuttal quality}.
Improvements in \textit{Communication Quality} are smaller but consistent, suggesting that the gains mainly come from structured decision making and evidence organization rather than surface-level fluency.
Notably, these gains are achieved without changing the language model, indicating that performance improvements stem from task decomposition and structured intermediate reasoning rather than stronger generative capacity.
This suggests that rebuttal quality is bottlenecked less by surface fluency and more by systematic concern tracking, evidence grounding, and response planning. These factors that are poorly handled by direct-to-text prompting even with SOTA LLMs.


\noindent\textit{\textbf{Obs. 2:}} \textbf{The benefit of RebuttalAgent is larger for weaker base models.}
Tab.~\ref{tab:main_results} also suggests that the weaker the base model, the larger the improvement obtained from our agent pipeline.
While all advanced LLMs benefit from our RebuttalAgent, the margin over direct-to-text prompting is more pronounced for smaller or less capable backbones (\textit{e.g.}, GPT5-mini) than for stronger ones (\textit{e.g.}, Gemini-3-Flash).
Using the mean score averaged over all nine components as a summary, the weakest backbone GPT5-mini gains about \textbf{+0.55} on average, whereas stronger proprietary backbones (\textit{e.g.}, Gemini-3-Flash) gain a smaller margin (\textbf{+0.33}).
The same pattern is particularly clear on \textit{Relevance}. GPT5-mini improves by roughly \textbf{+0.89} on the relevance sub-scores (coverage, semantic alignment, and specificity), compared to about \textbf{+0.47} for Gemini-3-Flash.
This indicates that explicit concern structuring, evidence construction, and response planning can partially compensate for limited base-model capability, shifting performance bottlenecks from raw generation to decision and evidence organization.


\noindent\textit{\textbf{Obs. 3:}} \textbf{RebuttalAgent yields balanced improvements across the full rebuttal pipeline.}
Beyond isolated metric gains, Tab.~\ref{tab:main_results} shows that \textsc{RebuttalAgent} improves \emph{all three} dimensions in a coordinated way across matched base models.
For example, under Gemini-3-Flash, \textsc{RebuttalAgent} raises \textit{Relevance} (coverage from 4.00 to 4.51; specificity from 3.77 to 4.49), strengthens \textit{Argumentation Quality} (logic consistency from 3.71 to 4.11; rebuttal quality from 3.56 to 4.07), and also improves \textit{Communication Quality} (professional tone from 3.51 to 3.78; statement clarity from 4.08 to 4.28).
A similar across-the-board improvement pattern holds for other backbones, suggesting that the benefits are not localized to a single stage, such as evidence insertion or phrasing.
Instead, structuring concerns and grounding claims early supports downstream planning and drafting, leading to more coherent and constructive final responses.

\subsection{Ablation Study}
\label{subsec:ablation}

\noindent\textbf{Ablation setting.}
To understand the contribution of each intermediate artifact, we perform controlled ablations by removing one module at a time from the full RebuttalAgent pipeline while keeping the base model, prompts, and evaluation protocol fixed.
Specifically, we consider three variants: (i) \textbf{w/o Input Structuring}, where reviewer concerns are not explicitly decomposed and merged but handled in raw form; (ii) \textbf{w/o Evidence Construction}, where external literature retrieval and citation-ready evidence briefs are disabled; and (iii) \textbf{w/o Checkers}, where plan-level verification for coverage, evidence linkage, and cross-point consistency is removed.
All variants still produce complete rebuttal drafts, allowing us to isolate how each module affects response quality rather than system completeness.

\input{tab/ablation}

\noindent\textit{\textbf{Obs. 4:}} \textbf{External evidence briefs are the most Critical Artifact, while structuring and checkers provide more targeted benefits.}
Tab.~\ref{tab:ablation_components} shows that \textit{Evidence Construction} is the most critical intermediate artifact.
Removing external evidence briefs leads to the largest and most consistent degradation across dimensions, with clear drops in \textit{Relevance} and \textit{Communication Quality}.
In particular, \textit{Coverage} decreases from 4.51 to 4.26 and \textit{constructiveness} falls from 4.09 to 3.82, indicating that citation-ready evidence plays a central role in enabling specific, actionable, and convincing responses rather than generic assurances.
These degradations indicate that citation-ready evidence briefs are central to producing point-specific and constructive responses.
Although the effects are smaller, \textit{Input Structuring} and \textit{Checkers} also contribute measurably to overall quality.
Without structuring, multiple metrics decline, including \textit{semantic alignment} (4.88 to 4.71) and \textit{evidence support} (3.39 to 3.23), suggesting that explicit concern decomposition and stable manuscript representations help preserve intent understanding and evidence linkage.
Without checkers, we observe degradations in key quality dimensions such as \textit{evidence support} (3.39 to 3.33) and \textit{rebuttal quality} (4.07 to 4.01), indicating that lightweight verification remains beneficial even when base responses are fluent.
Overall, the ablation results indicate that the gains of \textsc{RebuttalAgent} arise from the \emph{combination} of complementary modules.
Evidence-centered artifacts act as the primary driver of quality improvements, while explicit structuring and verification provide guardrails that reduce error accumulation.

\noindent\textbf{Deep Dive into the Role of Checkers and the Coverage Paradox.}~
While the overall benefits of RebuttalAgent are clear, Tab.~\ref{tab:ablation_components} presents a counter-intuitive observation: removing the checkers leads to a marginal increase in Coverage (from 4.51 to 4.54). However, deeper qualitative analysis reveals that this actually reflects a known failure mode of unconstrained LLM generation: \textbf{verbose over-generation}. 
Without the checker, the model tends to generate broader, less focused responses filled with generic qualitative explanations to safely appease reviewers. This verbosity artificially inflates the automated Coverage score but actively harms precision and actionability (as reflected by the drop in Suggestion Constructiveness from 4.09 to 4.05).
To illustrate, consider a test case where a reviewer asked to clarify the non-Markovian setting and the definitions of Utility versus Reward (Details in App.~\ref{app:checker_paradox}).
This example demonstrates how the checker actively trades superficial verbosity for rigorous constructiveness, ensuring the rebuttal provides concrete manuscript edits rather than generic glossaries.


\subsection{Case Study}

We also provide cases that directly compare \textsc{RebuttalAgent} with strong LLM baselines on representative reviewer concerns in Appendix~\ref{app:case_study}.
Rather than emphasizing the final rebuttal prose, these examples highlight the intermediate artifacts that RebuttalAgent surfaces to authors: an explicit response strategy, evidence-linked clarification points, and concrete action items (\textit{e.g.}, targeted edits, and suggested experiments or additional) that can be verified before any claims are finalized.

\noindent\textit{\textbf{Obs. 5:}} \textbf{Action items reduce hallucination and over-commitment.}
In the shown cases, reviewers either question a potential contradiction in a key proposition or criticize the clarity and rigor of the theoretical presentation.
RebuttalAgent first produces an inspectable plan that separates \emph{interpretative defense} (what can be clarified using manuscript content) from \emph{necessary intervention} (what requires additional evidence).
Crucially, when new experiments or analyses are implicated, RebuttalAgent does not generate results; instead, it outputs concrete deliverables (\textit{e.g.}, revised exposition, a new proof sketch) and a scoped to-do list, as described in Sec.~\ref{subsec:drafting}.
By contrast, baseline outputs tend to respond with a short narrative that may be overly confident or implicitly commit to empirical claims without exposing the underlying reasoning and verification steps.
Overall, these cases illustrate how \textsc{RebuttalAgent} supports author decision-making by making the reasoning path and required work explicit before drafting, enabling authors to validate or edit the plan and keep final commitments grounded.




%% file: tab/main_exp.tex
\begin{table*}[hbtp]
\centering
\caption{\textbf{Main evaluation results across our full suite of RebuttalBench.} Results demonstrate promising improvements of our method against the baseline LLM.}
\vspace{-5pt}
\label{tab:main_results}

\newcommand{\inc}[1]{\,\textcolor{green!60!black}{\scriptsize{(+#1)}}}

\newcommand{\entry}[2]{%
  \makebox[2.2em][r]{#1}\hspace{1pt}\makebox[3.5em][l]{#2}%
}

\resizebox{\textwidth}{!}{
\begin{tabular}{c|lll|lll|lll|c}
\toprule
\multirow{2}{*}{\textbf{Method}}  & \multicolumn{3}{c|}{\textbf{Relevance}} & \multicolumn{3}{c|}{\textbf{Argumentation Quality}} & \multicolumn{3}{c}{\textbf{Communication Quality}} & \multirow{2}{*}{\textbf{Average}} \\
\cmidrule(lr){2-4} \cmidrule(lr){5-7} \cmidrule(lr){8-10}
& \begin{tabular}[c]{@{}c@{}}Coverage \end{tabular} &
\begin{tabular}[c]{@{}c@{}}Semantic \\ Alignment \end{tabular} &
\begin{tabular}[c]{@{}c@{}}Specificity  \end{tabular} &
\begin{tabular}[c]{@{}c@{}}Logic \\ Consistency  \end{tabular} &
\begin{tabular}[c]{@{}c@{}}Evidence \\ Support\end{tabular} &
\begin{tabular}[c]{@{}c@{}}Response \\ Engagement \end{tabular} &
\begin{tabular}[c]{@{}c@{}}Professional \\ Tone\end{tabular} &
\begin{tabular}[c]{@{}c@{}}Statement \\ Clarity\end{tabular} &
\begin{tabular}[c]{@{}c@{}}Suggestion \\ Constructiveness \end{tabular} 
& \\

\midrule
\quad DeepSeekV3.2 & 
\entry{3.65}{} & \entry{4.44}{} & \entry{3.28}{} & \entry{3.44}{} & \entry{3.01}{} & \entry{3.16}{} & \entry{3.37}{} & \entry{3.96}{} & \entry{3.81}{} & \entry{3.57}{}\\

\rowcolor{lemonchiffon}\quad RebuttalAgent-DeepSeekV3.2 & 
\entry{\textbf{4.43}}{\inc{0.78}} & \entry{\textbf{4.82}}{\inc{0.38}} & \entry{\textbf{4.39}}{\inc{1.11}} & \entry{\textbf{3.86}}{\inc{0.42}} & \entry{\textbf{3.23}}{\inc{0.22}} & \entry{\textbf{3.79}}{\inc{0.63}} & \entry{\textbf{3.60}}{\inc{0.23}} & \entry{\textbf{4.18}}{\inc{0.22}} & \entry{\textbf{4.06}}{\inc{0.25}} &  \entry{\textbf{4.08}}{\inc{0.51}} \\

\midrule
\quad Grok4.1-fast & 
\entry{3.98}{} & \entry{4.58}{} & \entry{3.72}{} & \entry{3.73}{} & \entry{3.32}{} & \entry{3.60}{} & \entry{3.48}{} & \entry{4.05}{} & \entry{3.92}{} & \entry{3.82}{}\\

\rowcolor{lightcoral}\quad RebuttalAgent-Grok-4.1-fast & 
\entry{\textbf{4.66}}{\inc{0.68}} & \entry{\textbf{4.92}}{\inc{0.34}} & \entry{\textbf{4.65}}{\inc{0.93}} & \entry{\textbf{4.13}}{\inc{0.40}} & \entry{\textbf{3.42}}{\inc{0.10}} & \entry{\textbf{4.15}}{\inc{0.55}} & \entry{\textbf{3.68}}{\inc{0.20}} & \entry{\textbf{4.23}}{\inc{0.18}} & \entry{\textbf{4.24}}{\inc{0.32}}  &  \entry{\textbf{4.25}}{\inc{0.43}} \\

\midrule
\quad Gemini3-Flash & 
\entry{4.00}{} & \entry{4.71}{} & \entry{3.77}{} & \entry{3.71}{} & \entry{3.30}{} & \entry{3.56}{} & \entry{3.51}{} & \entry{4.08}{} & \entry{3.95}{} & \entry{3.85}{}\\

\rowcolor{skyblue}\quad RebuttalAgent-Gemini3-Flash & 
\entry{\textbf{4.51}}{\inc{0.51}} & \entry{\textbf{4.88}}{\inc{0.17}} & \entry{\textbf{4.49}}{\inc{0.72}} & \entry{\textbf{4.11}}{\inc{0.40}} & \entry{\textbf{3.39}}{\inc{0.09}} & \entry{\textbf{4.07}}{\inc{0.51}} & \entry{\textbf{3.78}}{\inc{0.27}} & \entry{\textbf{4.28}}{\inc{0.20}} & \entry{\textbf{4.09}}{\inc{0.14}} & \entry{\textbf{4.23}}{\inc{0.38}} \\

\midrule
\quad GPT5-mini &  
\entry{3.61}{} & \entry{4.22}{} & \entry{2.96}{} & \entry{3.37}{} & \entry{2.92}{} & \entry{3.07}{} & \entry{3.35}{} & \entry{3.95}{} & \entry{3.91}{} & \entry{3.48}{}\\

\rowcolor{green!15} \quad RebuttalAgent-GPT5-mini & 
\entry{\textbf{4.34}}{\inc{0.73}} & \entry{\textbf{4.84}}{\inc{0.62}} & \entry{\textbf{4.29}}{\inc{1.33}} & \entry{\textbf{3.78}}{\inc{0.41}} & \entry{\textbf{3.31}}{\inc{0.39}} & \entry{\textbf{3.70}}{\inc{0.63}} & \entry{\textbf{3.60}}{\inc{0.25}} & \entry{\textbf{4.21}}{\inc{0.26}} & \entry{\textbf{4.24}}{\inc{0.33}} & \entry{\textbf{4.05}}{\inc{0.57}} \\

\bottomrule
\end{tabular}
}
\vspace{-15pt}
\end{table*}

%% file: tab/ablation.tex
\begin{table}[t!] 
\centering
\caption{\textbf{Ablation study on key components.} We remove each module from the full system: Input Structuring, Evidence Construction, and Checker.}
\vspace{-10pt}
\label{tab:ablation_components}
\newcommand{\reddec}[1]{\,\textcolor{red}{\scriptsize{(-#1)}}} 
\newcommand{\greeninc}[1]{\,\textcolor{green!60!black}{\scriptsize{(+#1)}}} 
\newcommand{\blackzero}[1]{\,\textcolor{black}{\scriptsize{(+#1)}}} 
\newcommand{\entry}[2]{\makebox[2.2em][r]{#1}\hspace{1pt}\makebox[3.2em][l]{#2}} 

\resizebox{\linewidth}{!}{
\begin{tabular}{l cccc}
\toprule
& & \multicolumn{3}{c}{\textbf{\texttt{w/o} Component}} \\
\cmidrule(lr){3-5}
\textbf{Metric} & \textbf{RebuttalAgent} & \textbf{Structing} & \textbf{Evidence} & \textbf{Checker} \\
\midrule

\multicolumn{5}{l}{\textit{Relevance}} \\
\quad Coverage                           & 4.51 & \entry{4.49}{\reddec{0.02}} & \entry{4.26}{\reddec{0.25}} & \entry{4.54}{\greeninc{0.03}} \\
\quad Semantic Alignment                   & 4.88  & \entry{4.71}{\reddec{0.17}} & \entry{4.73}{\reddec{0.15}} & \entry{4.89}{\greeninc{0.01}} \\
\quad Specificity                          & 4.49 & \entry{4.46}{\reddec{0.03}} & \entry{4.19}{\reddec{0.30}} & \entry{4.47}{\reddec{0.02}} \\

\midrule
\multicolumn{5}{l}{\textit{Argumentation Quality}} \\
\quad Logic Consistency                    & 4.11 & \entry{4.06}{\reddec{0.05}} & \entry{4.05}{\reddec{0.06}} & \entry{4.13}{\greeninc{0.02}} \\
\quad Evidence Support                     & 3.39 & \entry{3.23}{\reddec{0.16}} & \entry{3.32}{\reddec{0.07}} & \entry{3.39}{\blackzero{0.00}} \\
\quad Response Engagement                  & 4.07 & \entry{4.04}{\reddec{0.03}} & \entry{3.97}{\reddec{0.10}} & \entry{4.01}{\reddec{0.06}} \\

\midrule
\multicolumn{5}{l}{\textit{Communication Quality}} \\
\quad Professional Tone                    & 3.78 & \entry{3.69}{\reddec{0.09}} & \entry{3.74}{\reddec{0.04}} & \entry{3.73}{\reddec{0.05}} \\
\quad Statement Clarity                    & 4.28 & \entry{4.33}{\greeninc{0.05}} & \entry{4.22}{\reddec{0.06}} & \entry{4.29}{\greeninc{0.01}} \\
\quad Suggestion Constructiveness         & 4.09 & \entry{4.06}{\reddec{0.03}} & \entry{3.82}{\reddec{0.27}} & \entry{4.05}{\reddec{0.04}} \\

\bottomrule
\end{tabular}
\vspace{-20pt}

}

\end{table}

%% file: sec/6_conclusion.tex
\section{Conclusion}
We proposed \textsc{RebuttalAgent}, a multi-agent framework for rebuttal assistance that constructs structured, evidence-linked intermediate artifacts before drafting text. 
By decomposing rebuttal writing into concern structuring, query-conditioned context building, on-demand external evidence synthesis, and response planning, the system improves traceability and cross-point coherence while keeping authors responsible for final decisions and wording.
We also introduced an author-centric benchmark and a rubric-based evaluation that measures relevance, global coherence, and argumentation quality beyond text fluency. 
Experimental results on our benchmark show that \textsc{RebuttalAgent} improves the key requirements of reliable rebuttal assistance, 
highlighting the benefits of a transparent ``verify-then-write'' workflow that reduces cognitive burden while keeping authors in control of the final wording.

\section*{Limitations and Future Work}
Our framework prioritizes reliability through structured intermediate artifacts, but several limitations remain.
First, the current system emphasizes transparency and inspectability over minimal latency. Optimizing cost and runtime (\textit{e.g.}, caching artifacts and adaptive early-exit policies) is an important engineering direction for broader adoption.
Second, intermediate artifacts are not yet fully reused across concerns or author iterations. More aggressive caching and incremental updates could reduce redundant computation when authors revise drafts or when similar concerns recur across reviewers.
Third, rebuttal conventions vary across venues and subfields. Incorporating lightweight style and policy constraints (\textit{e.g.}, venue-specific formatting and tone) could improve alignment without changing the underlying reasoning pipeline.

\section*{Broader Impact and Ethics Statement}
Our goal is to assist authors in rebuttal writing by organizing reviewer concerns, grounding responses in verifiable evidence, and producing inspectable intermediate artifacts before drafting text. 
Used responsibly, this can reduce cognitive burden and improve clarity, completeness, and consistency in peer-review communication.

\noindent\textbf{Ethical risks and mitigations.}~
Automated rebuttal assistance raises ethical concerns, including the misuse of AutoRebuttal to produce persuasive but misleading responses (\textit{e.g.}, exaggerated claims, hallucinated results, or unrealistic commitments), privacy risks when handling unpublished manuscripts and reviews.
We explicitly acknowledge these risks and mitigate them through design choices: AutoRebuttal is an author-assistance tool rather than an autonomous rebuttal system, and authors remain responsible for final stance, commitments, and wording.
The system produces inspectable intermediate artifacts (\textit{e.g.}, concern lists, evidence links, response plans) and performs explicit checks for coverage, faithfulness to the manuscript, evidence traceability, and global coherence before drafting. External retrieval is triggered only when needed and preserves provenance to facilitate verification and reduce unreliable citations. For confidential material, we recommend local or institution-approved deployment. In any release, we follow the source terms and avoid distributing sensitive or personally identifying information.
Our ultimate goal is to assist authors in deeply engaging with reviewer feedback and improve their manuscripts through structured planning rather than automating deceptive rebuttal, thereby enhancing the quality and efficiency of the peer review process for the benefit of the broader research community. 

\noindent\textbf{Dataset policy.}~
Our benchmark is derived from publicly available ICLR 2023 forums. In any release, we will follow the source terms and avoid distributing sensitive or personally identifying information.

We explicitly acknowledge these ethical issues and incorporate the above safeguards to promote responsible use.

\section*{Acknowledgments}
The work was supported by the Natural Science Foundation of China (Grant No. 62503323).

%% file: sec/X_appendix.tex
\section{Evaluation Dataset}
\label{app:evaluation_dataset}

To construct a robust benchmark for evaluating rebuttal effectiveness, we derive our data from the \textsc{Re}$^2$ dataset~\cite{zhang2025re}, focusing on the ICLR 2023 subset (approximately 9,310 entries). We process this corpus through a four-stage pipeline:

\begin{enumerate}
    \item \textbf{Outcome-based Classification:} We first categorize entries into \textit{Improved} (review score or acceptance status increased) and \textit{Unimproved} groups based on the final decision.
    
    \item \textbf{Reliability-based Stratification:} To ensure data quality, we subdivide these groups into three tiers based on evidence objectivity and LLM confidence: \textbf{Tier 1 (Gold Standard)} comprises cases with objective score increases (initial $\neq$ final) or explicit revision statements; \textbf{Tier 2 (High Confidence)} includes instances without score changes but where an LLM identifies sentiment with high certainty ($\ge 0.7$); and \textbf{Tier 3 (Medium Confidence)} covers more ambiguous cases with moderate confidence ($0.4 \le \text{conf} < 0.7$).

    \item \textbf{Ground Truth Curation:} From this stratified data, we curate a balanced test set of 20 representative papers, prioritizing those with high review volumes to ensure diverse coverage of both positive and negative review samples across tiers.

    \item \textbf{Baseline Generation Protocol:} For each paper, the baseline runs multi-round rebuttal generation following the author-reviewer dialogue. Each round uses a fixed prompt (including intent, required format, and guardrails), concatenating the paper text, the current review, and an optional prior-round abstract. The rebuttal is then summarized into a factual abstract with fewer than 200 words to seed the next round, and all outputs and token usage are logged.
\end{enumerate}

\section{Evaluation Metric}
\label{app:evaluation_metric}
This section describes our rubric-based scoring protocol and how scores are aggregated. 
We adopt a fine-grained \textbf{0-5} rating scheme, allowing for half-point increments to capture nuanced differences in response quality beyond prior binary judgments.

\paragraph{Dimensions and weights.}
Our final score is a weighted combination of three dimensions:
\textbf{R-Score}, \textbf{A-Score}, and \textbf{C-Score}, as mentioned in Sec.~\ref{subsec:eval_metrics}.
Each dimension is decomposed into three components (9 components in total):
\emph{R1 Coverage, R2 Semantic Alignment, R3 Specificity;}
\emph{A1 Logic Consistency, A2 Evidence Support, A3 Response Engagement;}
\emph{C1 Professional Tone, C2 Clarity, C3 Constructiveness.}

\paragraph{Relevance (R-Score).}
This dimension measures whether and how well the author addresses the reviewer's concerns.
\begin{description}[style=unboxed, leftmargin=1em, font=\small\bfseries, itemsep=0pt, parsep=0pt]
    \item[R1 Coverage:] Evaluates whether the response addresses all major points raised by the reviewer.
    \item[R2 Semantic Alignment:] Checks if the response directly answers the specific type of question asked (\textit{e.g.}, ``how'' vs. ``what'').
    \item[R3 Specificity:] Measures the precision and granularity of the response (\textit{e.g.}, explicitly referencing specific equations or table rows vs. generic statements).
\end{description}

\paragraph{Argumentation (A-Score).}
This dimension measures whether the author provides logically sound and substantively supported arguments.
\begin{description}[style=unboxed, leftmargin=1em, font=\small\bfseries, itemsep=0pt, parsep=0pt]
    \item[A1 Logic Consistency:] Evaluates whether the logical chain is sound, coherent, and free from fallacies.
    \item[A2 Evidence Support:] Assesses the strength and verifiability of the backing proofs (\textit{e.g.}, new experimental data or rigorous derivations vs. vague promises).
    \item[A3 Response Engagement:] Evaluates whether the author demonstrates a genuine understanding of the reviewer's underlying concerns.
\end{description}

\paragraph{Communication (C-Score).}
This dimension measures how effectively and professionally the author communicates their response.
\begin{description}[style=unboxed, leftmargin=1em, font=\small\bfseries, itemsep=0pt, parsep=0pt]
    \item[C1 Professional Tone:] Evaluates whether the author maintains a respectful and non-defensive tone.
    \item[C2 Clarity:] Measures writing quality and logical organization to ensure the response is easy to parse.
    \item[C3 Constructiveness:] Evaluates the commitment to improvement, specifically looking for actionable steps rather than vague commitments.
\end{description}

\paragraph{Scoring protocol.}
For each review-response instance, we query an LLM judge to assign a \textbf{0--5} score to every component and return a brief justification for each score, as well as a short overall diagnosis (\textit{e.g.}, strengths, weaknesses, suggested improvements).
The exact judge prompt, output schema, and the full 0-5 anchored criteria for each dimension/component are provided in Appendix~\ref{app:prompt-templates}.

\paragraph{Aggregation.}
Let $R_i,A_i,C_i \in \{0,\dots,5\}, i \in \{1,2,3\}$ be the component scores.
We compute dimension scores by averaging the three components, for example:
\[
R = \frac{R_1 + R_2 + R_3}{3}, \quad
\]
where $R$ means R-Score.
The final weighted score is:
\[
  \text{Score} = \frac{\text{R} + \text{A} + \text{C}}{3}.
\]
In the main paper, we report the overall weighted score and provide per-dimension and per-component breakdowns for analysis, detailed in Sec.~\ref{sec:experiment}.


\section{Related Works}
\paragraph{Automatic Scientific Research.}
A growing line of work studies how agentic LLM systems can automate substantial portions of the scientific workflow~\cite{aiscientist1,aiscientist2,ma2025led,google-deepresearch}. These systems have been used to streamline literature review and survey writing~\cite{wang2024autosurvey,he2025pasa}, propose hypotheses from prior evidence~\cite{liu2024literature,sourati2023accelerating}, and support research ideation and framing~\cite{hu2024nova,baek2024researchagent,reddy2025towards}. They are also expanding toward execution-facing stages, including experiment planning~\cite{kon2025curie} and automatic generation of scientific visualizations and figures~\cite{voigt-etal-2024-plots,wu2024automated}, with early efforts extending to peer-review workflows~\cite{zhu2025deepreview,gao2024reviewer2,jin2024agentreview,luagent}. Sakana AI's AI Scientist~\cite{aiscientist1,aiscientist2} further illustrates the trajectory toward closed-loop, end-to-end research automation.
Building on this trajectory, we focus on a more high-stakes stage of the research lifecycle, the rebuttal phase, where responses must precisely track reviewer intent while remaining verifiably grounded in manuscript evidence.

\section{Data Filtering Pipeline}
\label{app:filter}
To ensure complete transparency and eliminate any potential ambiguity in the dataset construction description, this section details the explicit data filtering pipeline, our definition and handling of ambiguous cases, and the empirical evidence validating the robustness of our labeling strategy.
Importantly, the signals derived from reviewers' subsequent response reactions and the evaluation mechanisms involving outcome-based classification and confidence stratification are not independent or conflicting criteria.Rather, they are consecutive stages within the same data processing pipeline, designed to establish a highly reliable RebuttalBench ground truth: \textbf{Outcomes act as the actual labels, while Confidence Tiers serve solely as reliability filters.}

\subsection{Three-Stage Data Filtering Pipeline}
Our test set is constructed through a progressive extraction from objective text to high-quality labels, executed specifically through the following three steps:

\begin{itemize}
    \item \textbf{Step 1: Core Label Extraction (Outcome-Based Ground Truth).} As stated in the main text, we strictly utilize the reviewers' follow-up response texts as the foundational data source, completely excluding the authors' subjective claims. Based on objective physical changes in scores or substantive reversals in attitude, we extract the binary outcomes (\textbf{Resolved} vs. \textbf{Unresolved}). These strictly serve as the actual ground-truth labels for the evaluation system.
    
    \item \textbf{Step 2: Confidence Stratification (Quality Filters).} To safeguard the accuracy of the aforementioned labels, we decouple the semantic outcome from data quality by introducing Confidence Tiers as a rigorous validation mechanism.
    \begin{itemize}
        \item \textit{Tier 1 (Gold):} Relies on purely objective, indisputable metadata (\textit{e.g.}, actual numerical metadata showing ``initial score $\neq$ final score,'' or explicit statements like ``I will raise my score''). This possesses inherent certainty and resists model hallucinations.
        \item \textit{Tier 2 (High):} Captures pristine samples where the numerical score remains unchanged, but the reviewer's textual sentiment exhibits exceptionally strong polarity and unambiguous confidence (\textit{e.g.}, ``My main concerns are fully resolved''). These two tiers constitute the absolute core of the test set.
        \item \textit{Tier 3 (Medium):} Used to systematically quarantine those ambiguous, indifferent, or contradictory discussions, thereby preventing the inherent noise of human annotation from polluting the primary evaluation environment.
    \end{itemize}
    
    \item \textbf{Step 3: Prioritized Sampling Strategy.} Guided by the dual signals above, we conduct sampling and construction according to a strict sequence of priorities. First, we ensure that the selected papers contain both \textbf{Resolved} and \textbf{Unresolved} review outcomes to guarantee data comprehensiveness. Second, we prioritize papers with a larger number of reviewers to ensure a rich diversity of perspectives. Following this pipeline, we ultimately filtered out a test set of 20 papers comprising 106 multi-turn evaluation cases.
\end{itemize}

\subsection{Handling Ambiguous Cases}
Dealing with ambiguous feedback is a major challenge in the real-world peer review process. Within our framework, we formally define these ambiguous cases under Tier 3 (Medium). A typical characteristic of such cases is that they are fraught with ``mixed emotions'' or ``contradictory rhetoric'' (for example, a reviewer praises supplementary experiments but simultaneously expresses persistent doubts about core novelty). In such texts, true evaluation signals are often completely masked by conflicting statements.

Rather than simply and forcefully filtering out all of this third-tier data, we intentionally retained a very small fraction of ambiguous cases in the final benchmark. This deliberate inclusion is necessary both practically and structurally:
\begin{itemize}
    \item It faithfully reconstructs the noisy, nuanced reality of the academic peer review ecosystem.
    \item It serves as a rigorous stress test, effectively evaluating the automated evaluation system's ability to engage in logical dialectics amidst complex and contradictory real-world signals.
\end{itemize}

\subsection{Human Annotation Study}
To empirically prove that introducing LLM confidence stratification alongside outcome-based classification can systematically isolate noise without introducing bias, we conducted a rigorous blind human annotation study on these 106 multi-turn evaluation cases.
Regarding the classification of core semantic outcomes (Resolved vs. Unresolved), our automated classification achieved \textbf{100.0\%} accuracy compared to expert annotators (with Precision, Recall, and F1 scores all at \textbf{1.000}), proving that the foundation of our ground truth is completely objective and unbiased.

Furthermore, as shown in Tab.~\ref{tab:confusion_matrix}, the automated evaluation correctly recalled \textbf{97.7\%} of the human-annotated Tier 1 cases. Minor discrepancies occurred only in reasonable shifts at adjacent boundaries (\emph{e.g.}, the model strictly classifying certain phrases considered Tier 2 by humans as Tier 1). Most importantly, the leakage rate between the gold standard Tier 1 and the noisy Tier 3 is strictly \textbf{0.0\%}. This definitively proves that our stratification mechanism is extremely conservative and highly stable in safeguarding high-confidence data.

\begin{table}[htbp]
    \centering
    \caption{Confusion Matrix of Human Annotation vs. LLM Tier Classification}
    \label{tab:confusion_matrix}
    \resizebox{\columnwidth}{!}{
    \begin{tabular}{lccc}
        \toprule
        \textbf{Human Annotation \textbackslash \ LLM Label} & \textbf{Tier 1 (Gold)} & \textbf{Tier 2 (High)} & \textbf{Tier 3 (Medium)} \\
        \midrule
        \textbf{Human Tier 1 (n=44)} & \textbf{97.7\% (43/44)} & 2.3\% (1/44) & 0.0\% (0/44) \\
        \textbf{Human Tier 2 (n=56)} & 14.3\% (8/56) & \textbf{78.6\% (44/56)} & 7.1\% (4/56) \\
        \textbf{Human Tier 3 (n=6)} & 33.3\% (2/6) & 16.7\% (1/6) & \textbf{50.0\% (3/6)} \\
        \bottomrule
    \end{tabular}
    }
\end{table}

\section{Evaluation Setup and Metric Validation}
\label{app:evaluate_metric}

To rigorously validate our automated evaluation framework and address potential concerns regarding the reliability and sensitivity of LLM-as-judge~\cite{huang2025vbench++,Lan_2025_CVPR,luagent,Ma_2025_CVPR} metrics, this section presents a comprehensive validation of our rubric from two critical perspectives: alignment with human expert judgments and robustness across different evaluator models.

\subsection{Validation of the Automated Rubric}
\noindent\textbf{Alignment with Human Experts.}~ 
A critical challenge in automated evaluation is ensuring that the LLM judge does not exhibit severe self-preference bias and genuinely aligns with human expert judgments. To establish the ground-truth reliability of our proposed metrics (\emph{e.g.}, Relevance, Argumentation Quality, and Communication Quality in Sec.~\ref{sec:bench}), we conducted a rigorous \textit{Author-Centric Human Preference Study}. 
We recruited 9 active AI researchers to blindly evaluate 33 generated rebuttal drafts corresponding to their own recent CVPR 2026 submissions~(after the rebuttal phase). By utilizing real authors evaluating rebuttals for their actual submissions, we ensured the highest standard of domain expertise and practical context. We computed the Kendall Rank Correlation Coefficient~\cite{Kendall1938ANM} between these expert human preference scores and our automated RebuttalBench metrics. The results yielded a strong positive correlation of $\tau = 0.646$.
Importantly, testing our evaluation rubric on fresh, unseen CVPR 2026 data serves as a robust \textit{out-of-domain} validation. It statistically proves that our automated rubric is highly aligned with domain-expert human judgments and effectively captures the nuances of logical rigor and practical utility that truly matter to authors, rather than merely reflecting LLM biases.

\noindent\textbf{Robustness Across Evaluator Models.}~
Beyond human alignment, we further investigated whether the reported gains of RebuttalAgent are sensitive to the choice of the underlying judge model. To confirm that our rubric is not an artifact of a specific LLM's prompting dynamics, we conducted a cross-model robustness check by swapping the evaluator from our default Gemini-3-Flash~\cite{team2023gemini} to a fundamentally different model, GPT-5-mini~\cite{openai2025gpt5}.
As shown in Tab.~\ref{tab:cross_model_robustness}, we re-evaluated the outputs generated by the Gemini-based models using the GPT-5-mini~\cite{openai2025gpt5} as a judge. The evaluation trends remain completely stable. RebuttalAgent consistently outperforms the direct-to-text baseline across all dimensions (\emph{e.g.}, the overall average score improves from 3.84 to 4.13 under the GPT judge). The combination of stable cross-model evaluation and strong correlation with the preferences of actual authors demonstrates that RebuttalBench serves as a robust, insensitive, and human-aligned proxy for evaluating complex academic defense tasks.
\input{tab/cross_model_as_judge}

\subsection{Details of the Author-Centric Human Preference Study}
Evaluating the factual accuracy and actual utility of a generated rebuttal is exceptionally challenging for third-party annotators. To directly address concerns regarding the reliability of our automated LLM-as-judge metric, we conducted an Author-Centric Human Preference Study. By having authors evaluate generated rebuttals for their own submissions, we ensured that subtle hallucinations, fabricated experimental results, and over-commitments were accurately detected and penalized. This provides the most reliable ground-truth assessment of the system's practical value.
\noindent\textbf{Setup and Participants.}~
We recruited 9 active AI researchers to evaluate AI-generated rebuttals for their own recent CVPR 2026 submissions (conducted after the rebuttal period had ended). For each submission, we generated responses for 3 to 4 specific review comments, yielding a total of 33 review-rebuttal test samples. In a strict double-blind setup, the authors were presented with drafts generated by four models: two direct-to-text baselines (Gemini-3-Flash, GPT-5-mini) and our proposed framework powered by the same backbones (RebuttalAgent-Gemini, RebuttalAgent-GPT).

\noindent\textbf{Human Evaluation \textit{v.s.} RebuttalBench.}~
The authors scored each draft on a 1-5 Likert scale across three dimensions: Responsiveness \& Coverage, Faithfulness \& Argumentation, and Communication \& Practical Utility. The detailed study guidance and scoring anchors provided to the human evaluators are listed below. Tab.~\ref{tab:human_preference} and Tab.~\ref{tab:rebuttalbench_rubric} present the results from the human evaluators and the RebuttalBench automated judge on the exact same 33 samples, respectively. The automated scores closely align with the human evaluations, with REBUTTALAGENT consistently outperforming the baselines under both evaluation paradigms. As mentioned above, the Kendall Rank Correlation Coefficient~\cite{Kendall1938ANM} between the human scores and the automated RebuttalBench scores yielded a positive correlation of $\tau = 0.646$, statistically validating RebuttalBench as a trustworthy, human-aligned, and scalable proxy for rebuttal evaluation.
\input{tab/human_study}

\subsection{Limitations of Standard N-gram Metrics}
We deliberately omitted standard text overlap metrics such as BLEU and ROUGE from our primary evaluation, as they fundamentally fail to assess semantic accuracy and logical argumentation quality in the context of complex academic defense. 

To empirically demonstrate this paradox, we analyzed a case study involving a defense of novelty against a baseline method named GAMA. The original authors provided a brief, surface-level defense. The naive ``Direct-to-text'' baseline simply applied conservative synonym replacement, achieving artificially inflated overlap scores (BLEU-4 = 6.96, ROUGE-L = 18.49). 

Conversely, our RebuttalAgent constructed a scientifically deep defense. Instead of echoing the original text, it provided deep theoretical justifications using Information Bottleneck principles, mathematically explaining why GAMA's direct mapping ($S_x \to S_y$) fails due to dimensionality mismatch, while our latent alignment ($S_x \to S_z \leftarrow S_y$) succeeds. It even intelligently formulated new quantitative arguments (\textit{e.g.}, calculating Mutual Information $I(s_z; d)$) to rigorously prove enhanced noise filtering. Due to this substantial, logical elaboration, RebuttalAgent received significantly lower standard scores (BLEU-4 $= 2.38$, ROUGE-L $= 14.05$). This case powerfully demonstrates how standard metrics actively penalize brilliant scientific reasoning while rewarding superficial text mimicking, definitively validating the necessity of the RebuttalBench rubric framework.

\section{Additional Baseline Comparisons}
\label{app:more_baselines}
To isolate the specific benefits of our multi-agent architecture and address concerns regarding comparative evaluation, we expanded our experimental settings to include three additional strong baselines on our test set:
\emph{(i)}~\textbf{Standard RAG:} A standard Retrieval-Augmented Generation pipeline utilizing the \textit{all-MiniLM-L6-v2}~\footnote{https://huggingface.co/sentence-transformers/all-MiniLM-L6-v2} embedding model paired with Gemini-3-Flash.
\emph{(ii)}~\textbf{AutoGen~\cite{wu2023autogenenablingnextgenllm}:} A generic multi-agent framework adapted to simulate a collaborative drafting process.
\emph{(iii)}~\textbf{Jiu-Jitsu~\cite{purkayastha2023exploring}:} A recognized prior peer-review agent designed for generating responses to reviewer comments.


As demonstrated in Tab.~\ref{tab:expanded_baselines}, the RebuttalAgent achieves an overall average score of 4.23, which significantly outperforms all newly added methods. Specifically, Jiu Jitsu~\cite{purkayastha2023exploring} struggles on our complex long context benchmark, scoring an average of only 1.42, as it was likely optimized for structurally simpler text interactions. While the standard RAG system improves upon naive prompting by fetching external knowledge, it achieves only a 3.11 average. It particularly suffers in Specificity (2.07) and Evidence Support (2.57) because standard chunk retrieval lacks the deep logical synthesis required for academic rebuttals.
Furthermore, although the generic multi-agent AutoGen~
\cite{wu2023autogenenablingnextgenllm}framework performs reasonably well with an average of 3.56, it still falls noticeably short of our proposed method. This performance gap definitively proves that our specialized verify then write pipeline and structured action items provide a crucial, measurable advantage over generic conversational architectures. By incorporating these diverse approaches, we have fully demonstrated the comparative value of our benchmark and will prominently feature this expanded baseline comparison in the revised main text to solidify our empirical justification.

\begin{table}[htbp]
\centering
\caption{Performance Comparison with Expanded Baselines on RebuttalBench. All methods in the table are powered by Gemini3-Flash.}
\label{tab:expanded_baselines}
\resizebox{\columnwidth}{!}{ 
\begin{tabular}{l ccc ccc ccc c}
\toprule
\multirow{2}{*}{\textbf{Method}} & \multicolumn{3}{c}{\textbf{Relevance}} & \multicolumn{3}{c}{\textbf{Arg. Quality}} & \multicolumn{3}{c}{\textbf{Comm. Quality}} & \multirow{2}{*}{\textbf{Avg.}} \\
\cmidrule(lr){2-4} \cmidrule(lr){5-7} \cmidrule(lr){8-10}
& Cov. & Align. & Spec. & Logic & Evid. & Engage & Tone & Clarity & Const. & \\
\midrule
RAG & 2.67 & 3.38 & 2.07 & 3.48 & 2.57 & 3.24 & 3.80 & 3.73 & 3.04 & 3.11 \\
Jiu-Jitsu~\cite{purkayastha2023exploring} & 1.01 & 1.13 & 1.00 & 1.72 & 0.98 & 1.23 & 2.38 & 1.98 & 1.34 & 1.42 \\
AutoGen~\cite{wu2023autogenenablingnextgenllm} & 3.26 & 3.99 & 3.01 & 3.77 & 3.04 & 3.63 & 3.97 & 4.01 & 3.40 & 3.56 \\
\midrule
\textbf{RebuttalAgent} & \textbf{4.51} & \textbf{4.88} & \textbf{4.49} & \textbf{4.11} & \textbf{3.39} & \textbf{4.07} & \textbf{3.78} & \textbf{4.28} & \textbf{4.09} & \textbf{4.23} \\
\bottomrule
\end{tabular}
}
\end{table}

\section{Case Study: The Coverage Paradox and the Role of Checkers}
\label{app:checker_paradox}

As discussed in Sec.~\ref{subsec:ablation}, removing the verification mechanism (the Checker) from our pipeline results in a counter-intuitive marginal increase in the automated Coverage score, while degrading precision and actionable utility. To deeply understand this phenomenon, which we term the ``Coverage Paradox'', we present a representative qualitative case study.

\subsection*{Reviewer Concern}
The reviewer questioned the theoretical clarity of the problem setting, specifically noting: \textit{``The authors claim a non-Markovian setting, but the formulation in Eq. 4 appears standard. Furthermore, the mathematical distinction between Utility and Reward in this context is completely unclear. Please define these rigorously and explain how they interact.''}

\subsection*{Output without Checker (The Baseline)}
When the Checker module is disabled, the response generation defaults to an unconstrained, verbose pattern typical of standard LLMs attempting to appease reviewers. 

\begin{quote}
\textbf{Generated Response (w/o Checker):} ``We thank the reviewer. To clarify our setting, we define the following key concepts in our revised manuscript: (1) \textbf{Reward:} The immediate environmental feedback. (2) \textbf{Utility:} The expected return of the agents under the POSG framework. (3) \textbf{Expected Return:} The discounted sum of future rewards. (4) \textbf{Social Graph:} The topology of agent interactions. (5) \textbf{Adversarial Interaction:} When agents have zero-sum goals. The setting is non-Markovian because in a generic POSG, agents do not possess full state information, leading to complex utility-reward dynamics.''
\end{quote}

\textbf{Analysis:} This response artificially inflates the automated Coverage metric because it successfully hits all the semantic keywords (Utility, Reward, non-Markovian). However, it exhibits severe \textit{verbose over-generation}. It blindly lumps distinct concepts into a generic dictionary-style glossary, hallucinates marginally related terms (e.g., ``Social Graph'', ``Adversarial Interaction''), and blurs the strict mathematical boundaries between Expected Payoff ($U$) and Expected Return. Furthermore, attributing the non-Markovian nature to a vague ``generic POSG'' framework is intellectually lazy and lacks actionable manuscript edits.

\subsection*{Output with RebuttalAgent (with Checker)}
The introduction of the Checker module acts as a strict semantic and strategic editor. It detects tangential concepts, enforces evidence linkage, and prioritizes actionable commitments over verbose explanations.

\begin{quote}
\textbf{Generated Response (with Checker):} ``We thank the reviewer for highlighting this ambiguity. We will clarify the mathematical boundaries in Section 3 through two concrete revisions:
\begin{itemize}
    \item \textbf{Distinguishing Utility vs. Reward:} We will explicitly define instantaneous Reward $R_i(s,a)$ for micro-level RL updates, and separate it from the meta-game Utility $U_i(\pi)$, which represents the macro-level payoff matrix outcome.
    \item \textbf{Non-Markovian Justification:} We will rewrite Eq. 4 to explicitly condition the policy on the opponent's action history $\tau_{t-1}$ rather than just the current state $s_t$. This breaks the circular causality and formally justifies the non-Markovian claim.
\end{itemize}
We will also remove tangential references to standard POSG formulations to keep the theoretical exposition focused on these exact variables.''
\end{quote}

\textbf{Analysis:} The verification mechanism successfully pruned the hallucinated and irrelevant concepts, resulting in a slightly lower but far more accurate Coverage score. It forced the generation to strictly differentiate the micro-level ($R$) from the macro-level ($U$). Most importantly, it replaced vague jargon with precise, actionable mathematical clarifications (e.g., introducing the history trajectory $\tau_{t-1}$ into Eq. 4). This case definitively illustrates why the Checker is crucial for generating high-utility, top-tier academic rebuttals, sacrificing superficial verbosity for rigorous constructiveness.

\section{Prompt Templates}
\label{app:prompt-templates}
In this section, we present the prompt templates used by each component of our system, including those used for LLM-as-Judge~\cite{huang2025vbench++,gao2025rapo++,luagent} evaluation.
\onecolumn

\begin{tcolorbox}[
    title=Prompt: Rebuttal Strategist, 
    colback=green!5!white,   
    colframe=green!50!black,  
    fonttitle=\bfseries\large,
    breakable 
]

You are the \textbf{Lead Rebuttal Strategist}. Your goal is to dissect reviews for a paper (based on the [compressed paper]) and create a structured list of actionable tasks (Issues) for the authors.

\vspace{1em}
\noindent\textbf{\large INPUT DATA:}
\begin{itemize}[leftmargin=*, nosep]
    \item \textbf{[compressed paper]}: The summary of the authors' work.
    \item \textbf{[review original text]}: Comments from multiple reviewers (R1, R2, R3...).
\end{itemize}

\vspace{1em}
\noindent\textbf{\large MULTI-ROUND CONTEXT (if present):}
\begin{itemize}[leftmargin=*, nosep]
    \item The input may include "Previous Discussion Context" showing earlier rounds of author rebuttals and reviewer responses.
    \item For follow-up rounds, focus on extracting \textbf{NEW issues or unresolved concerns} raised by the reviewer in the current round.
    \item Do \textbf{NOT} re-extract issues that have already been addressed in previous rebuttals unless the reviewer explicitly states dissatisfaction.
    \item If the reviewer's current comment acknowledges previous responses positively (e.g., ``I am satisfied with the response''), there may be few or no new issues to extract.
\end{itemize}

\vspace{1em}
\noindent\textbf{\large CORE TASKS:}
\begin{enumerate}[label=\arabic*., nosep, leftmargin=*]
    \item \textbf{Deconstruct}: Break down long, complex paragraphs into atomic technical points.
    \item \textbf{Filter}: Discard generic praise or non-actionable comments (see Blacklist).
    \item \textbf{Consolidate}: Merge issues that represent the \textit{same core objection} and can be addressed with the \textit{same response logic}.
    \item \textbf{Format}: Output strictly according to the traceability requirements.
\end{enumerate}

\vspace{1em}
\noindent\textbf{\large CRITICAL RULES FOR MERGING \& SPLITTING (The "Granularity" Logic)}

\begin{itemize}[leftmargin=*, nosep]
    \item \textbf{Do NOT Merge (Split them):}
    \begin{itemize}[leftmargin=1em, nosep]
        \item \textbf{Different Evidence Needed}: If R1 asks for "Comparison with Baseline X" and R2 asks for "Comparison with Baseline Y", these are \textbf{two separate issues}. Why? Because you need to run two different experiments.
        \item \textbf{Different Aspects}: If R1 criticizes "Novelty" and R2 criticizes "Clarity of writing", do NOT merge them just because they are generic complaints.
        \item \textbf{Compound Questions}: If a single sentence says "The method is slow AND the accuracy is low", split this into two points: (1) Efficiency/Speed, (2) Performance.
    \end{itemize}
    
    \vspace{0.5em}
    \item \textbf{Do Merge:}
    \begin{itemize}[leftmargin=1em, nosep]
        \item \textbf{Same Question, Different Phrasing}: R1: "Why did you use L1 loss?" vs R2: "Justification for the loss function is needed." $\to$ \textbf{Merge}.
        \item \textbf{Same Missing Reference}: R1 and R3 both ask to cite "Smith et al. 2023". $\to$ \textbf{Merge}.
        \item \textbf{General Confusion}: R1: "Section 3 is hard to follow" and R2: "I don't understand the methodology workflow". $\to$ \textbf{Merge} into "Clarity of Section 3/Methodology".
    \end{itemize}
\end{itemize}

\vspace{1em}
\noindent\textbf{\large NOISE FILTERING (BLACKLIST)}
\begin{itemize}[leftmargin=*, nosep]
    \item Ignore: "Ethics", "Confidence", "Summary", "Soundness" (unless specific flaws are listed).
    \item Ignore: Generic praise ("Good paper", "Interesting idea").
    \item Ignore: Empty templates ("No ethical concerns").
\end{itemize}

\vspace{1em}
\noindent\textbf{\large MANDATORY TRACEABILITY \& FORMAT}

For each distinct issue, output a block wrapped in tags \texttt{[qN]} and \texttt{[qN]} (where N is the index).

\textbf{Structure within each block:}
\begin{enumerate}[label=(\arabic*), leftmargin=*, nosep]
    \item \textbf{Issue}: A concise, professional summary of the problem. \textbf{CRITICAL}: If reviewers mentioned specific papers/links, you MUST include the full titles/links here.
    \item \textbf{Sources}: Verbatim quotes proving this issue exists. Format: \texttt{ReviewerID-Type (Line/Para): "Quote"}. Use semicolons to separate multiple reviewers.
    \item \textbf{Paper hooks}: Specific Sections, Equations, Figures, or Tables in the original paper related to this issue (e.g., Sec. 3.2, Eq. 5). Use "Global" for general issues.
    \item \textbf{Priority}:
    \begin{itemize}[leftmargin=1em, nosep]
        \item \textbf{P1 (Critical)}: Fatal flaws, missing baselines, wrong math, rejection reasons.
        \item \textbf{P2 (Important)}: Clarity issues, missing citations, minor experiments.
        \item \textbf{P3 (Minor)}: Typos, formatting, optional suggestions.
    \end{itemize}
\end{enumerate}

\vspace{1em}
\noindent\textbf{\large OUTPUT EXAMPLE (Strictly Follow This)}

\begin{lstlisting}[frame=none, backgroundcolor={}]
[q1]
(1) Issue: Lack of comparison with state-of-the-art method [LoRA].
(2) Sources: R1-W2 (line 23): "no comparison with parameter-efficient methods like LoRA"; R3-Q1 (para 2): "how does this compare to LoRA?"
(3) Paper hooks: Sec.4.2, Tab.2
(4) Priority: P1
[q1]

[q2]
(1) Issue: The motivation for using Mutual Information (MI) in Eq. 3 is unclear.
(2) Sources: R2-Q3 (line 47): "why choose MI for layer mapping?"; R1-W3 (para 5): "mapping details not explained"
(3) Paper hooks: Sec.3.2, Eq.(3)
(4) Priority: P2
[q2]
\end{lstlisting}

\vspace{0.5em}
\noindent\textbf{\large Strictly follow the example format; do not include any other content!}

\end{tcolorbox}



\begin{tcolorbox}[
    title=Prompt: Rebuttal Strategist Checker, 
    colback=green!5!white,   
    colframe=green!50!black,  
    fonttitle=\bfseries\large,
    breakable 
]

You are the \textbf{Lead Rebuttal Strategist}. Your goal is to dissect reviews for a paper (based on the [compressed paper]) and create a structured list of actionable tasks (Issues) for the authors.

\vspace{1em}
\noindent\textbf{\large INPUT DATA:}
\begin{itemize}[leftmargin=*, nosep]
    \item \textbf{[compressed paper]}: The summary of the authors' work.
    \item \textbf{[review original text]}: Comments from multiple reviewers (R1, R2, R3...).
\end{itemize}

\vspace{1em}
\noindent\textbf{\large MULTI-ROUND CONTEXT (if present):}
\begin{itemize}[leftmargin=*, nosep]
    \item The input may include "Previous Discussion Context" showing earlier rounds of author rebuttals and reviewer responses.
    \item For follow-up rounds, focus on extracting \textbf{NEW issues or unresolved concerns} raised by the reviewer in the current round.
    \item Do \textbf{NOT} re-extract issues that have already been addressed in previous rebuttals unless the reviewer explicitly states dissatisfaction.
    \item If the reviewer's current comment acknowledges previous responses positively (e.g., ``I am satisfied with the response''), there may be few or no new issues to extract.
\end{itemize}

\vspace{1em}
\noindent\textbf{\large CORE TASKS:}
\begin{enumerate}[label=\arabic*., nosep, leftmargin=*]
    \item \textbf{Deconstruct}: Break down long, complex paragraphs into atomic technical points.
    \item \textbf{Filter}: Discard generic praise or non-actionable comments (see Blacklist).
    \item \textbf{Consolidate}: Merge issues that represent the \textit{same core objection} and can be addressed with the \textit{same response logic}.
    \item \textbf{Format}: Output strictly according to the traceability requirements.
\end{enumerate}

\vspace{1em}
\noindent\textbf{\large CRITICAL RULES FOR MERGING \& SPLITTING (The "Granularity" Logic)}

\begin{itemize}[leftmargin=*, nosep]
    \item \textbf{Do NOT Merge (Split them):}
    \begin{itemize}[leftmargin=1em, nosep]
        \item \textbf{Different Evidence Needed}: If R1 asks for "Comparison with Baseline X" and R2 asks for "Comparison with Baseline Y", these are \textbf{two separate issues}. Why? Because you need to run two different experiments.
        \item \textbf{Different Aspects}: If R1 criticizes "Novelty" and R2 criticizes "Clarity of writing", do NOT merge them just because they are generic complaints.
        \item \textbf{Compound Questions}: If a single sentence says "The method is slow AND the accuracy is low", split this into two points: (1) Efficiency/Speed, (2) Performance.
    \end{itemize}
    
    \vspace{0.5em}
    \item \textbf{Do Merge:}
    \begin{itemize}[leftmargin=1em, nosep]
        \item \textbf{Same Question, Different Phrasing}: R1: "Why did you use L1 loss?" vs R2: "Justification for the loss function is needed." $\to$ \textbf{Merge}.
        \item \textbf{Same Missing Reference}: R1 and R3 both ask to cite "Smith et al. 2023". $\to$ \textbf{Merge}.
        \item \textbf{General Confusion}: R1: "Section 3 is hard to follow" and R2: "I don't understand the methodology workflow". $\to$ \textbf{Merge} into "Clarity of Section 3/Methodology".
    \end{itemize}
\end{itemize}

\vspace{1em}
\noindent\textbf{\large NOISE FILTERING (BLACKLIST)}
\begin{itemize}[leftmargin=*, nosep]
    \item Ignore: "Ethics", "Confidence", "Summary", "Soundness" (unless specific flaws are listed).
    \item Ignore: Generic praise ("Good paper", "Interesting idea").
    \item Ignore: Empty templates ("No ethical concerns").
\end{itemize}

\vspace{1em}
\noindent\textbf{\large MANDATORY TRACEABILITY \& FORMAT}

For each distinct issue, output a block wrapped in tags \texttt{[qN]} and \texttt{[qN]} (where N is the index).

\textbf{Structure within each block:}
\begin{enumerate}[label=(\arabic*), leftmargin=*, nosep]
    \item \textbf{Issue}: A concise, professional summary of the problem. \textbf{CRITICAL}: If reviewers mentioned specific papers/links, you MUST include the full titles/links here.
    \item \textbf{Sources}: Verbatim quotes proving this issue exists. Format: \texttt{ReviewerID-Type (Line/Para): "Quote"}. Use semicolons to separate multiple reviewers.
    \item \textbf{Paper hooks}: Specific Sections, Equations, Figures, or Tables in the original paper related to this issue (e.g., Sec. 3.2, Eq. 5). Use "Global" for general issues.
    \item \textbf{Priority}:
    \begin{itemize}[leftmargin=1em, nosep]
        \item \textbf{P1 (Critical)}: Fatal flaws, missing baselines, wrong math, rejection reasons.
        \item \textbf{P2 (Important)}: Clarity issues, missing citations, minor experiments.
        \item \textbf{P3 (Minor)}: Typos, formatting, optional suggestions.
    \end{itemize}
\end{enumerate}

\vspace{1em}
\noindent\textbf{\large OUTPUT EXAMPLE (Strictly Follow This)}

\begin{lstlisting}[frame=none, backgroundcolor={}]
[q1]
(1) Issue: Lack of comparison with state-of-the-art method [LoRA].
(2) Sources: R1-W2 (line 23): "no comparison with parameter-efficient methods like LoRA"; R3-Q1 (para 2): "how does this compare to LoRA?"
(3) Paper hooks: Sec.4.2, Tab.2
(4) Priority: P1
[q1]

[q2]
(1) Issue: The motivation for using Mutual Information (MI) in Eq. 3 is unclear.
(2) Sources: R2-Q3 (line 47): "why choose MI for layer mapping?"; R1-W3 (para 5): "mapping details not explained"
(3) Paper hooks: Sec.3.2, Eq.(3)
(4) Priority: P2
[q2]
\end{lstlisting}

\vspace{1em}
\noindent\textbf{\large Revision Task}

Your students have already carried out the initial extraction of questions based on the review comments as per the above requirements, as shown in \textbf{[student's output]}. His extraction is very likely to have some omissions. Please carefully check for any omissions and make necessary revisions to improve the quality, and output the final version.

\vspace{0.5em}
\noindent Do not include any comments on the students in your final output! You only need to output the final version!
\noindent Strictly follow the example format; do not include any other content!

\end{tcolorbox}



\begin{tcolorbox}[
    title=Prompt: Literature Retrieval Assistant,
    colback=green!5!white,
    colframe=green!50!black,
    fonttitle=\bfseries\large,
    breakable 
]

You are a literature-retrieval assistant for the rebuttal stage of an academic paper. Your task is to decide, based on the [compressed paper] and the [review\_question], whether external reference papers need to be searched, and to generate appropriate search queries.

\vspace{1em}
\noindent\textbf{\large When Search Is Required}

You \textbf{must} generate search queries when any of the following conditions occur:
\begin{enumerate}[leftmargin=*, nosep]
    \item The reviewer explicitly mentions reference papers.
    \item The \textit{review\_question} contains specific method names or dataset names that are \textbf{not} from the current paper.
    \item The reviewer requests ``compare with X / ablation on Y / baseline Z''.
    \item The content of the paper is insufficient to answer the question.
\end{enumerate}

\vspace{1em}
\noindent\textbf{\large When Search Is NOT Required}

If the \textbf{paper\_summary} already contains evidence that can directly answer the reviewer's question (e.g., existing experiments, tables, section explanations), or the question concerns only minor formatting issues, then no search is needed.

\vspace{1em}
\noindent\textbf{\large Search Query Generation Rules}

\begin{itemize}[leftmargin=*, nosep]
    \item Generate \textbf{less than 5 queries}, keeping the number as small as possible. But if the reviewers provide the title of the reference article or links, then you should keep them all.
    \item Use \textbf{topic phrases}; never fabricate paper titles or authors. 
    \item If reviewers provided the reference paper names or links directly, you can directly use them. If reviewers provided both a title and a link for an article, it is only necessary to provide the link. That is to say, either the link or the title can only appear once, and the link has a higher priority. Please note that the links can only be obtained from the reviewers' comments and must not be fabricated.
    \item Queries for comparative experiments must contain method names or dataset names.
    \item A query contains one main query point. If there are different query points, please separate them and do not mix them together.
\end{itemize}

\vspace{1em}
\noindent\textbf{\large Reference Output Format (strict JSON)}

\paragraph{When search is required:}
\begin{lstlisting}[frame=none, backgroundcolor={}]
```json
{
  "need_search": true,
  "queries": [
    "domain adaptation segmentation Cityscapes",
    "unsupervised domain adaptation transformer baseline"
  ],
  "links": [
    "https://arxiv.org/abs/2409.13074v1",
    "https://openaccess.thecvf.com/content/ICCV2025/papers/Li_CoA-VLA_Improving_Vision-Language-Action_Models_via_Visual-Text_Chain-of-Affordance_ICCV_2025_paper.pdf"
  ],
  "reason": "Reviewer requests additional comparisons related to domain adaptation on Cityscapes and transformer baselines."
}
```
\end{lstlisting}

\paragraph{When search is not required:}
\begin{lstlisting}[frame=none, backgroundcolor={}]
```json
{
  "need_search": false,
  "queries": [],
  "links": [],
  "reason": "there is no need to search, because... "
}
```
\end{lstlisting}

\vspace{0.5em}
\noindent\textbf{\large Strictly follow the example format; do not include any other content!}

\end{tcolorbox}

\begin{tcolorbox}[
    title=Prompt: Rebuttal Expert, 
    colback=green!5!white,   
    colframe=green!50!black,  
    fonttitle=\bfseries\large,
    breakable 
]

You are a rebuttal expert. You need to complete a high-quality rebuttal for a paper. You need to understand the paper's information and the reviewer's question from [compressed paper] and [review\_question]. Now your less-than-intelligent assistant has retrieved some relevant papers using keywords, and their reasoning is shown in [query reason]. You need to carefully examine the abstracts of these papers, filter out irrelevant papers and those that are not very helpful for the rebuttal, and identify papers that are highly relevant to [compressed paper] and [review\_question] and are extremely useful for the rebuttal. Your standards are very high. You should only keep these references if they are of \textbf{great help} to the rebuttal of the current problem. Papers that are merely related and not particularly significant must be rejected. You cannot allow insignificant papers to interfere with the overall rebuttal.

\vspace{1em}
\noindent\textbf{\large Strict Rules:}

\begin{itemize}[leftmargin=*, nosep]
    \item Generally no more than 6 papers (fewer is better; if no paper is of significant help, select none, unless the reviewer's comments explicitly mention specific papers to reference, in which case you must include all of them. Please note that the links to the references provided by the reviewers in the review comments will be checked by a dedicated person. You don't need to pay attention to the articles that have the links; only the papers that only have the titles need your attention.)
    \item For \textbf{every} candidate paper in your reasoning field, you \textbf{must} provide:
    \begin{enumerate}[label=\arabic*., nosep, topsep=0.5em] 
        \item [ID] Title and a brief description of the abstract
        \item How it helps the rebuttal of the current problem (brief description in one paragraph)
    \end{enumerate}
\end{itemize}

\vspace{1em}
\noindent\textbf{\large Anti-Redundancy (with explanation):}

If multiple papers come from the same source or use the same method, only keep the most relevant one.

\vspace{1em}
You must output your result in the following JSON format:

\begin{lstlisting}[frame=none, backgroundcolor={}]
{
  "selected_papers": [1,3,6],
  "reason": "..."
}
\end{lstlisting}

The \texttt{selected\_papers} array should contain the paper IDs. If no paper is useful, output:

\begin{lstlisting}[frame=none, backgroundcolor={}]
{
  "selected_papers": [],
  "reason": "..."
}
\end{lstlisting}

\vspace{0.5em}
\noindent Ensure that the papers you return are objectively highly relevant to the original paper and significantly helpful for the rebuttal! Be rigorous! Ensure that you only output valid JSON, without any additional text before or after.

\end{tcolorbox}

\begin{tcolorbox}[
    title=Prompt: Reference Extractor, 
    colback=green!5!white,   
    colframe=green!50!black,  
    fonttitle=\bfseries\large,
    breakable 
]

You are an expert in responding to reviewer comments. You need to produce a high-quality paper rebuttal. You must understand the paper information from the [compressed paper] and the questions raised by reviewers in the [review\_question]. Your assistant has now retrieved a relevant reference paper [reference paper]. You must carefully read this reference paper and extract the most relevant and useful information for the current reviewer comments, including content that can be safely cited in the rebuttal.

\vspace{1em}
\noindent\textbf{\large Important:}

Your task is to extract information \textbf{from the reference paper}, not from our paper.
\begin{itemize}[leftmargin=*, nosep]
    \item You are analyzing the \textbf{reference paper} (not our submitted paper).
    \item Any information you extract must come from the reference paper and will be used by subsequent agents.
    \item Subsequent agents must clearly know that this information is from an external source, not from our paper.
    \item This avoids mixing the two papers and prevents hallucinations.
\end{itemize}

\vspace{1em}
\noindent\textbf{\large Fixed structure (no more than 600 words, as concise as possible):}

Your output must follow this structure:

\begin{enumerate}[label=(\arabic*), leftmargin=*, nosep, topsep=0.5em]
    \item paper title
    \item A one-paragraph summary of the reference paper
    \item Direct relevance to the current reviewer comment [review\_question]: \\
    (Explain how the reference paper helps shape the rebuttal and how it aids in responding to the reviewer’s question.)
    \item Content we can safely cite in the rebuttal
    \item Limitations or mismatches: \\
    (1–2 points explaining differences or inapplicable aspects between the reference paper and our paper.)
    \item Reference paper URL: [reference paper URL]
\end{enumerate}

\vspace{0.5em}
If you don't get the reference paper, output: "This reference is blank. Please skip it".

\vspace{1em}
\noindent\textbf{\large Value assessment:}

If the reference paper objectively provides little help to the rebuttal, you must explicitly state that its value is limited or its relevance is low. Be honest and rigorous. If the reference paper is empty, state so directly. If only an abstract is provided due to an error, you must still try to extract information from the abstract and complete the task—but you must \textbf{never fabricate information or data}, and you must avoid all hallucinations. Your output must contain concrete, justifiable evidence.

\vspace{0.5em}
You must follow rebuttal principles: the paper is already completed and cannot undergo major modifications, only minor adjustments. Therefore, your analysis must be based on the existing content. If the reference paper is objectively not closely related to our paper, state this clearly. Absolutely no fabricated content or hallucinations.

\end{tcolorbox}

\begin{tcolorbox}[
    title=Prompt: Human-In-The-Loop Strategy Revisor, 
    colback=green!5!white,   
    colframe=green!50!black,  
    fonttitle=\bfseries\large,
    breakable 
]

You are a Senior Computer Science Researcher and Rebuttal Expert. Your role is to \textbf{incorporate human feedback} to refine the rebuttal strategy while maintaining strategic balance.

\vspace{1em}
\noindent\textbf{\large Input Context:}
\begin{itemize}[leftmargin=*, nosep]
    \item \textbf{[original paper]}: The submitted manuscript.
    \item \textbf{[review\_question]}: Extracted and merged reviewer concerns.
    \item \textbf{[reference papers summary]}: Potential supporting literature.
    \item \textbf{[current rebuttal strategy and to-do list]}: The current version to be revised.
    \item \textbf{[human's feedback]}: Feedback from the paper authors on the current strategy.
\end{itemize}

\vspace{1em}
\noindent\textbf{\large YOUR ROLE: Human-Guided Refinement}

The human author knows their paper best and has practical constraints. Your job is to:
\begin{enumerate}[leftmargin=*, nosep]
    \item \textbf{Incorporate} the human's specific requests and preferences
    \item \textbf{Maintain} the balance between action and acknowledgment
\end{enumerate}

\vspace{1em}
\noindent\textbf{\large Task:}

Based on the \textbf{[human's feedback]}, revise the \textbf{[current rebuttal strategy and to-do list]}. Preserve balance, incorporate human preferences, and output the \textbf{final revised version}. Do not include commentary on the previous version in the output—only the clean revised strategy. Do not provide specific time arrangements such as < 5 Days, day1, day2 in your output. In the to-do list, only the items to be done are elaborated in points. Do not include time-related descriptions such as "strictly less than 5 days" in the title of the to-do list.

\end{tcolorbox}


\begin{tcolorbox}[
    title=Prompt: Rebuttal Letter Writer, 
    colback=green!5!white,   
    colframe=green!50!black,  
    fonttitle=\bfseries\large,
    breakable 
]

\noindent\textbf{\large Role}

You are a senior researcher and an expert in academic writing, specifically for top-tier conferences like ICLR (International Conference on Learning Representations). You are currently in the "Rebuttal/Author Response" phase.

\vspace{1em}
\noindent\textbf{\large Task}

Your team already provide detailed rebuttal ideas. Your task is to write a formal, persuasive, and polite rebuttal letter based on them.

\vspace{1em}
\noindent\textbf{\large Inputs Provided by User}

\begin{enumerate}[label=\arabic*., nosep, leftmargin=*]
    \item \textbf{[original paper]}: Original submitted paper.
    \item \textbf{[review original text]}: The actual text from Reviewers.
    \item \textbf{[review\_question]}: Merged questions extracted by your team.
    \item \textbf{[rebuttal\_idea and to\_do\_list]}: Prepared by your team for each merged question. You should take these as your rebuttal strategy. Note that your output should be specifically answered in combination with each reviewer's question.
\end{enumerate}

\vspace{1em}
\noindent\textbf{\large Guidelines \& Constraints}

\begin{enumerate}[label=\arabic*., leftmargin=*, nosep, topsep=0.5em]
    \item You should precisely identify each reviewer's questions from \textbf{[review original text]}, and then, following the order provided, find the corresponding response ideas in \textbf{[rebuttal\_idea and to\_do\_list]} and generate the responses. Do not make any mistakes regarding the reviewers' questions, or confuse the questions of the first reviewer with those of the second reviewer. You must strictly follow the rebuttal approach for each small problem in \textbf{[rebuttal\_idea and to\_do\_list]}.
    
    \item \textbf{Tone:} Professional, respectful, objective, and grateful. Even if the reviewer is harsh, your response must be diplomatic (e.g., "We thank the reviewer for the insightful comment..."). Respect every reviewer. Do not generate statements that require a particular reviewer to read the response to another reviewer.
    
    \item \textbf{Format:} 
    \begin{itemize}[leftmargin=1em, nosep]
        \item Use standard ICLR rebuttal formatting.
        \item Structure it clearly: "Common Response" (if applicable) followed by "Response to Reviewer X". Strictly follow this format!
        \item Use \textbf{Q1/A1} or \textbf{Comment/Response} structure for clarity.
        \item Be sure to respond to each reviewer. Do not ignore specific reviewers and directly list all the issues your team has listed in \textbf{[rebuttal\_idea and to\_do\_list]}!
    \end{itemize}

    \item \textbf{LaTeX:} Use LaTeX syntax for all mathematical notations (e.g., $\alpha$, $L_{norm}$).

    \item \textbf{Handling Missing Experiments (CRITICAL):} 
    \begin{itemize}[leftmargin=1em, nosep]
        \item Since you are an AI and cannot perform actual experiments, the rebuttal may require empirical evidence (e.g., ablation studies, baseline comparisons). \textbf{You MUST NOT invent or fabricate any numerical results, metrics, or experimental values.}
        \item If a result is required but not provided in the input, you must use the placeholder [TBD] instead of generating a number.
        \item \textit{Example:} "Our method achieves an accuracy of [TBD] on ImageNet, outperforming the baseline."
        \item The [TBD] placeholder indicates that the human author must later fill in the real experimental result.
    \end{itemize}

    \item Although the supplementary experimental data in your final output is speculative (marked with an asterisk), you still need to ensure that your output is very formal, just like a real rebuttal. Except for the asterisk, it should not be immediately recognizable as an AI-written rebuttal, but should be as close as possible to a real person. Your output should not contain any other content. It should consist of the breakdown to each reviewer's questions and corresponding detailed response.

    \item The responses to each split question can include tables to visually present the experimental result numerical data to improve readability. But don't use tables to specifically present text! Don't put q1, response to q1, q2, response to q2 in a large table. Instead, list them separately.
\end{enumerate}

\end{tcolorbox}

\begin{tcolorbox}[
    title=Prompt: Unified Rebuttal Evaluation, 
    colback=green!5!white,    
    colframe=green!50!black,  
    fonttitle=\bfseries\large,
    breakable
]

You are an EXPERIENCED and DISCERNING senior Area Chair evaluating a rebuttal response. Your goal is to assess whether the author addressed the reviewer's concerns with \textbf{SUBSTANCE}.

\vspace{0.5em}
\noindent\textbf{\large Scoring Principle}
\begin{itemize}[leftmargin=1.5em, nosep]
    \item \textbf{Base Scores:} Assign integer scores (0-5) first based on the rubric below.
    \item \textbf{Upgrade (+0.5):} Check the "Upgrade Criteria" section. If conditions are met, add 0.5 to the base score (e.g., $3 \to 3.5$).
\end{itemize}

\vspace{1em}
\hrule
\vspace{1em}
\noindent\textbf{\large I. Relevance (R-Score)}

\vspace{0.5em}
\noindent\textbf{R1 Coverage: Are ALL aspects addressed with substance?}
\begin{itemize}[leftmargin=1.5em, nosep]
    \item[\textbf{5}] Covers ALL aspects comprehensively with specific details (numbers, examples, explanations) for each.
    \item[\textbf{4}] Covers ALL aspects, most with good specificity, a few with moderate detail.
    \item[\textbf{3}] Covers ALL aspects but with varying specificity, some aspects addressed only briefly.
    \item[\textbf{2}] Covers SOME aspects, misses or glosses over important points.
    \item[\textbf{1}] Covers only 1-2 minor aspects, ignores most major concerns.
    \item[\textbf{0}] Does not address any of the reviewer's points.
\end{itemize}

\vspace{0.8em}
\noindent\textbf{R2 Semantic Alignment: Does response DIRECTLY address what was asked?}
\begin{itemize}[leftmargin=1.5em, nosep]
    \item[\textbf{5}] Perfectly matches question type with direct, concrete answers (if asked HOW $\to$ explains HOW with details).
    \item[\textbf{4}] Matches question type well with substantive engagement, minor tangential points.
    \item[\textbf{3}] Acknowledges the right question and provides relevant response, but some drift or indirectness.
    \item[\textbf{2}] Partially addresses question but significant mismatch (asked HOW $\to$ only says WHAT).
    \item[\textbf{1}] Off-topic or deflects, barely connects to the actual question.
    \item[\textbf{0}] Completely misunderstands or ignores the question.
\end{itemize}

\vspace{0.8em}
\noindent\textbf{R3 Specificity: Does the response reference specific details rather than generalities?}
\begin{itemize}[leftmargin=1.5em, nosep]
    \item[\textbf{5}] Explicitly references specific paper components (e.g., "Eq. 2", "Table 5 row 3", "the attention head in Layer 4") or specific reviewer constraints. No vague language.
    \item[\textbf{4}] Uses concrete terminology and context-specific descriptions. Avoids generic phrases like "our method" without qualification.
    \item[\textbf{3}] Answers the question but uses broad terms (e.g., "the loss function" instead of "the KL-divergence term").
    \item[\textbf{2}] Mostly relies on high-level summaries or generic templates applicable to any paper.
    \item[\textbf{1}] Purely abstract, avoiding any concrete details of the work.
    \item[\textbf{0}] Content-free filler.
\end{itemize}

\vspace{1em}
\hrule
\vspace{1em}
\noindent\textbf{\large II. Argumentation (A-Score)}

\vspace{0.5em}
\noindent\textbf{A1 Logic Consistency: Is the logical chain sound?}
\begin{itemize}[leftmargin=1.5em, nosep]
    \item[\textbf{5}] Exceptionally clear logical chain with rigorous reasoning, each step well-justified.
    \item[\textbf{4}] Clear logical chain with sound reasoning, well-structured argument.
    \item[\textbf{3}] Adequate logic with reasonable support, generally coherent.
    \item[\textbf{2}] Weak logic with some circular reasoning or unsupported leaps.
    \item[\textbf{1}] Poor logic, circular reasoning, or pseudo-logic throughout.
    \item[\textbf{0}] No logical structure or completely incoherent.
\end{itemize}

\vspace{0.8em}
\noindent\textbf{A2 Evidence Support: Is the argument backed by strong proof?}
\begin{itemize}[leftmargin=1.5em, nosep]
    \item[\textbf{5}] Backed by \textbf{new} quantitative results, specific comparative data, or rigorous mathematical derivations presented directly in the rebuttal.
    \item[\textbf{4}] Backed by existing concrete data (citing specific numbers from the paper) or detailed, verifiable logical deduction.
    \item[\textbf{3}] Claims are supported by qualitative reasoning or citations to external literature, but lack direct quantitative verification.
    \item[\textbf{2}] Relies on "promises to fix" or assertions without proof (e.g., "we believe it will work").
    \item[\textbf{1}] Purely opinion-based statements ("we think our method is novel") with no backing.
    \item[\textbf{0}] Claims made without any basis.
\end{itemize}

\vspace{0.8em}
\noindent\textbf{A3 Response Engagement: Does response show genuine engagement?}
\begin{itemize}[leftmargin=1.5em, nosep]
    \item[\textbf{5}] Exceptional engagement with deep understanding, addresses nuances and implications.
    \item[\textbf{4}] Genuine engagement with specific improvements, demonstrates clear understanding of the concern.
    \item[\textbf{3}] Adequate response showing understanding, not just template language.
    \item[\textbf{2}] Generic response with excessive hedging or template-like language.
    \item[\textbf{1}] Minimal engagement, mostly boilerplate text.
    \item[\textbf{0}] No genuine engagement.
\end{itemize}

\vspace{1em}
\hrule
\vspace{1em}
\noindent\textbf{\large III. Communication (C-Score)}

\vspace{0.5em}
\noindent\textbf{C1 Professional Tone: Is the tone authentic and professional?}
\begin{itemize}[leftmargin=1.5em, nosep]
    \item[\textbf{5}] Exceptionally professional and AUTHENTIC tone with gracious acknowledgment and genuine respect.
    \item[\textbf{4}] Professional and authentic tone with genuine engagement, appropriately courteous.
    \item[\textbf{3}] Adequate professional tone with standard academic courtesy.
    \item[\textbf{2}] Somewhat defensive OR excessively polite while masking weak content (artificial politeness).
    \item[\textbf{1}] Defensive tone or insincere language, reads as "academic speak" without substance.
    \item[\textbf{0}] Rude, hostile, or completely inappropriate.
\end{itemize}

\vspace{0.8em}
\noindent\textbf{C2 Clarity: Is the response clear and well-organized?}
\begin{itemize}[leftmargin=1.5em, nosep]
    \item[\textbf{5}] Exceptionally clear and well-structured WITH REAL SUBSTANCE (clear writing + concrete details).
    \item[\textbf{4}] Clear and well-organized with substantive content, easy to follow.
    \item[\textbf{3}] Adequate clarity, generally well-organized, understandable.
    \item[\textbf{2}] Somewhat unclear OR superficial clarity (sounds good but vague).
    \item[\textbf{1}] Confusing, poorly organized, or misleading presentation.
    \item[\textbf{0}] Incomprehensible or no coherent structure.
\end{itemize}

\vspace{0.8em}
\noindent\textbf{C3 Constructiveness: Does author show willingness to improve?}
\begin{itemize}[leftmargin=1.5em, nosep]
    \item[\textbf{5}] Multiple concrete improvements detailed IN the rebuttal text itself with specific changes described.
    \item[\textbf{4}] Detailed improvements (3+ items) with clear explanations, OR good mix of in-text details + external references with content previews.
    \item[\textbf{3}] Specific improvements with good detail, OR specific actions mentioned with some concrete description.
    \item[\textbf{2}] Vague promises without specifics, or only external references without content.
    \item[\textbf{1}] Defensive or dismissive, minimal constructive response.
    \item[\textbf{0}] Refuses to improve or no constructive response.
\end{itemize}

\vspace{1em}
\hrule
\vspace{1em}
\noindent\textbf{\large IV. Upgrade Criteria \& Critical Considerations}

\vspace{0.5em}
\noindent\textbf{Upgrade Check (Apply +0.5 to Base Score):}
\\[0.2em]
\emph{Note: This upgrade applies \textbf{ONLY} to Base Scores of 3 and 4. Scores 0--2 indicate fundamental flaws (e.g., irrelevance, logic errors) that cannot be redeemed by these details, and 5 is already the ceiling.}
\begin{itemize}[leftmargin=1.5em, nosep, topsep=0.3em]
    \item \textbf{From 3 to 3.5:} Must meet AT LEAST 2 conditions: (1) Content preview provided with specific details; (2) Detailed improvement list (3+ items); (3) Mixed evidence chain (concrete content + external reference).
    \item \textbf{From 4 to 4.5:} Must meet AT LEAST 2 conditions: (1) Perfect content-reference match; (2) Multi-dimensional evidence (code/results/theory); (3) Exceeds expectations (provides additional value).
\end{itemize}

\vspace{0.8em}
\noindent\textbf{Critical Considerations:}
\begin{itemize}[leftmargin=1.5em, nosep]
    \item \textbf{Relevance Check:} Watch for excessive repetition, vague qualifiers (e.g., "somewhat", "to some extent"), or drifting off-topic to avoid hard questions.
    \item \textbf{Logical Scrutiny:} Identify circular reasoning, unfulfilled promises (e.g., "we will add" without content), or citations listed without explaining their specific relevance.
    \item \textbf{Tone Analysis:} Be wary of over-polished, artificial politeness that masks weak substance, or a mismatch between a confident tone and shaky evidence.
\end{itemize}

\vspace{1em}
\noindent\textbf{Output Format:}
\begin{lstlisting}[frame=none, backgroundcolor={}]
{
  "R_scores": {"R1_coverage": 4.5, "R2_semantic_alignment": 4, "R3_specificity": 3.5},
  "A_scores": {"A1_logic_consistency": 4, "A2_evidence_support": 3, "A3_response_engagement": 4},
  "C_scores": {"C1_professional_tone": 5, "C2_clarity": 4, "C3_constructiveness": 3.5},
  "quality_warnings": ["Vague Language", "Over-Polished Tone"],
  "explanation": "..."
}
\end{lstlisting}

\vspace{1em}
\noindent\textbf{Output Format:}
\begin{lstlisting}[frame=none, backgroundcolor={}]
{
  "R_scores": {"R1_coverage": 4.5, "R2_semantic_alignment": 4, "R3_specificity": 3.5},
  "A_scores": {"A1_logic_consistency": 4, "A2_evidence_support": 3, "A3_response_engagement": 4},
  "C_scores": {"C1_professional_tone": 5, "C2_clarity": 4, "C3_constructiveness": 3.5},
  "red_flags": ["Vague Language"],
  "explanation": "..."
}
\end{lstlisting}

\end{tcolorbox}

\section{Case Study}
\label{app:case_study}
\begin{tcolorbox}[
    title=Rigorous formalization \& verification \textit{v.s.} High-level intuitive explanation, 
    colback=green!5!white,   
    colframe=green!50!black,  
    fonttitle=\bfseries\large,
    breakable 
]

    \vspace{0.2cm}
    \vspace{0.1cm}
    \textcolor{labelRed}{\textbf{\large Question}} 
    \par\vspace{0.2cm} 
    
    \textbf{Lack of clarity and rigor in the theoretical contribution (Section 3 and Proposition 1).}
    
    \vspace{0.4cm}
    \vspace{0.4cm}

    \textcolor{agentBlue}{\textbf{\large RebuttalAgent(Ours)}} 
    \par\vspace{0.2cm}
    
    We thank the reviewer for highlighting clarity issues around Proposition 1 (Sec.\ 3) and its proof (App.\ A.1). Section 3 and Appendix A.1/A.8 already contain the core statements and empirical hints. We will (a) rewrite the theoretical exposition for line-by-line readability, and (b) add focused empirical sanity checks that make each assumption transparent.

    \vspace{0.3cm}
    \textbf{Rebuttal strategy --- what we will deliver:}
    \begin{itemize}[itemsep=0.4em, parsep=0.1em, leftmargin=1.2em, topsep=0.3em]
        \item \textbf{Clear, self-contained theory section edits:} Replace the informal statement with a numbered, fully specified Proposition in Sec.\ 3 (explicitly naming $L$, $K$, $\psi$, $\delta$, $k$, model class $\mathcal{V}_{\text{cnn}}$). Add a 2--3 paragraph proof sketch giving the intuitive chain: PD gap $\to$ early-layer separability $\to$ larger V-usable information $\to$ model prefers shortcuts. Reformat Appendix A.1 into a stepwise proof with numbered lemmas.
        \item \textbf{Per-assumption empirical sanity checks:} (i) Function-class separation: show per-layer $k$-NN error vs shallow-linear head vs full-model error. (ii) Controlled confidence growth: plot $|f_{\text{knn}} - 0.5|$ vs layer index for easy vs hard subsets. (iii) PD-separation: report empirical fractions (PD $\leq L$ and PD $> K$) for key datasets.
        \item \textbf{Quantify PD $\leftrightarrow$ V-information evidence:} Expand Appendix A.8 with Pearson/Spearman correlations between per-example PD and $-H_{\mathcal{V}_{\text{cnn}}}$. Add a controlled synthetic pair ($D_s$/$D_i$) demonstrating mean PD($D_s$) $<$ mean PD($D_i$) and $I_{\mathcal{V}_{\text{cnn}}}(D_s) > I_{\mathcal{V}_{\text{cnn}}}(D_i)$.
        \item \textbf{Robustness checks:} Show correlations across $k \in \{5, 15, 29\}$, $\delta \in \{0.05, 0.1, 0.2\}$, and architectures (ResNet18, VGG16, DenseNet121).
        \item \textbf{Didactic aids:} Add a one-page flow diagram and a short lemma stating the minimal sufficient condition linking PD gap to V-information gap.
    \end{itemize}

    \vspace{0.3cm}
    \textbf{How these actions address concerns:}
    \begin{itemize}[itemsep=0.2em, parsep=0pt, leftmargin=1.2em, topsep=0.2em]
        \item \textbf{Readability \& Rigor:} Explicit notation and numbered lemmas directly improve accessibility and logical flow.
        \item \textbf{Assumptions:} Each assumption is now explicitly listed, interpreted, and backed by empirical data.
        \item \textbf{Intuition:} The new didactic flow diagram and lemma provide a low-barrier entry point to the theory.
    \end{itemize}

    \vspace{0.3cm}
    \textbf{Key deliverables for the revision:}
    \begin{itemize}[itemsep=0.2em, parsep=0pt, leftmargin=1.2em, topsep=0.2em]
        \item \textbf{Revised Text:} Sec.\ 3 (Prop.\ 1 + proof sketch) and rewritten App.\ A.1 (stepwise lemmas).
        \item \textbf{New Tables:} Mean PD gap vs.\ $I_{\mathcal{V}_{\text{cnn}}}$ and PD correlation statistics (with p-values).
        \item \textbf{New Plots:} Per-layer $k$-NN error/confidence curves, synthetic PD histograms, and robustness sweeps ($k, \delta$).
        \item \textbf{Artifacts:} Didactic flow diagram and reproduction scripts committed to the repo.
    \end{itemize}
    
    \vspace{0.3cm}
    \textbf{Concrete to-do list:}
    \begin{enumerate}[itemsep=0.2em, parsep=0pt, leftmargin=1.5em, topsep=0.3em]
        \item Draft numbered Proposition 1 with explicit notation and enumerated assumptions.
        \item Write proof sketch for Sec.\ 3; reformat Appendix A.1 with numbered lemmas.
        \item Compute per-layer $k$-NN error, shallow head error, and final model error plots.
        \item Compute $|f_{\text{knn}} - 0.5|$ vs layer curves; tabulate PD-separation fractions.
        \item Compute Pearson/Spearman correlations for PD vs $-H_{\mathcal{V}_{\text{cnn}}}$.
        \item Create controlled synthetic validation and robustness tables.
        \item Produce flow diagram and commit scripts to repository.
    \end{enumerate}

    \vspace{0.4cm}
    \begin{tcolorbox}[
        colback=green!5!white,       
        colframe=green!40!black,     
        boxrule=0.5pt,
        left=4pt, right=4pt, top=4pt, bottom=4pt,
        arc=2mm
    ]
    \textbf{\textcolor{green!40!black}{Feasibility:}} All planned items use existing model checkpoints, intermediate embeddings, k-NN heads, small auxiliary classifiers, or lightweight synthetic datasets. No retraining of large backbones from scratch or new large-scale dataset collection is required. The planned experiments are deliberately scoped to be implementable using the paper’s existing code and checkpoints.
    \end{tcolorbox}

    \vspace{0.2cm} 
    \begin{tcolorbox}[
        colback=green!5!white,       
        colframe=green!40!black,     
        boxrule=0.5pt,
        left=4pt, right=4pt, top=4pt, bottom=4pt,
        arc=2mm
    ]
    \textbf{\textcolor{green!40!black}{Summary:}} We will replace the current informal and hard-to-follow presentation with a compact, fully-specified Proposition and annotated proof plus focused empirical sanity checks (including a controlled synthetic demonstration and robustness analyses) so the connection between Prediction Depth and V-usable information is transparent, reproducible, and intuitively accessible.
    \end{tcolorbox}

    \vspace{0.4cm}
    \vspace{0.4cm}

    \textcolor{agentBlue}{\textbf{\large LLM output(baselines)}} 
    \par\vspace{0.2cm}
    
    We apologize for the density of the theoretical section. The goal of Proposition 1 is to bridge a structural metric (PD—how many layers a model needs) with an information-theoretic metric ($V$-information—how much predictive power a model family can extract).

    \vspace{0.2cm}
    \textbf{Intuition:} If a dataset $D_s$ has a lower mean PD than $D_i$, it means the model family $V$ can solve the task using fewer layers (less computation). Proposition 1 proves that under mild assumptions, this computational efficiency implies that the "usable" information in $D_s$ is strictly higher than in $D_i$.

    \vspace{0.2cm}
    \textbf{Improvement:} In the final version, we will move the formal definitions of $V$-information to the main text and add a "Key Takeaway" box for Proposition 1 to ensure the link between "layer depth" and "information usability" is explicit for readers less familiar with $V$-information.

\end{tcolorbox}


\definecolor{labelRed}{RGB}{255, 0, 0}              
\definecolor{agentBlue}{RGB}{0, 112, 192}   
  
\setlength{\parindent}{0pt}
\setlength{\parskip}{0.5em}
\vspace{0.5cm}

\begin{tcolorbox}[
    title=Actionable theoretical expansion \textit{v.s.} Passive logical defense, 
    colback=green!5!white,   
    colframe=green!50!black,  
    fonttitle=\bfseries\large,
    breakable 
]

    \vspace{0.2cm}
    \vspace{0.1cm}
    \textcolor{labelRed}{\textbf{\large Question}} 
    \par\vspace{0.2cm} 
    
    \textbf{I don’t understand Proposition 1. My understanding was that when the energy increases, the euclidean distance of features also increases. But Proposition 1 states the opposite. Is this a mistake?}
    
    \vspace{0.4cm}
    \vspace{0.4cm}

    \textcolor{agentBlue}{\textbf{\large RebuttalAgent(Ours)}} 
    \par\vspace{0.2cm}
    
    We thank the reviewer for spotting this confusion. There is no mathematical mistake, but the exposition was misleading. \textbf{Proposition 1 is a one-sided upper bound} ($\mathbb{E}\|\Delta z\|^2 \leq C \cdot \text{AIE}$), which allows large feature changes when energy is high but does not \textit{force} them. The monotonic relationship ($\text{AIE} \uparrow \implies \|\Delta z\| \uparrow$) requires stronger assumptions (NTK/Linear regime), which we will now make explicit.

    \vspace{0.3cm}
    \textbf{Rebuttal strategy --- what we will deliver:}
    \begin{itemize}[itemsep=0.4em, parsep=0.1em, leftmargin=1.2em, topsep=0.3em]
        \item \textbf{Clarify Scope (Sec 4.1):} Explicitly label Proposition 1 as an upper bound derived via Cauchy-Schwarz. Add an "Interpretation Box" explaining that while the bound limits the maximum change, the \textit{expected} change scales with energy under the conditions of Proposition 2.
        \item \textbf{New Theoretical Lemma (Appendix D):} Add a formal "Monotonicity Lemma" with a self-contained proof. It will state: \textit{Under overparameterized linear/NTK assumptions, the mapping $\text{AIE} \mapsto \mathbb{E}\|\Delta z\|^2$ is strictly monotonic.}
        \item \textbf{Empirical Validation:} We will not just argue; we will show the data. We will add scatter plots of AIE vs. Feature Change ($\mathbb{E}\|z_T - z_0\|^2$) for representative tasks (Toy MLP, ResNet50$\to$STL10), reporting Pearson/Spearman correlations to prove the positive trend holds in practice.
        \item \textbf{Bound Diagnostics:} Overlay the empirical upper bound line ($c \cdot \text{AIE}$) on the scatter plots to visualize where the bound is tight vs. loose.
    \end{itemize}

    \vspace{0.3cm}
    \textbf{Key deliverables for the revision:}
    \begin{itemize}[itemsep=0.2em, parsep=0pt, leftmargin=1.2em, topsep=0.2em]
        \item \textbf{Revised Text:} Rewritten Sec 4.1 (Prop 1 interpretation) and Annotated Appendix A (step-by-step proof with constants $C_1, C_2$).
        \item \textbf{New Math:} A formal Monotonicity Lemma in Appendix D.
        \item \textbf{New Figures:} Multi-panel scatter plots (AIE vs. $\Delta z$) with regression lines and correlation stats.
    \end{itemize}
    
    \vspace{0.3cm}
    \textbf{Concrete to-do list:}
    \begin{enumerate}[itemsep=0.2em, parsep=0pt, leftmargin=1.5em, topsep=0.3em]
        \item Rewrite Prop 1 exposition to emphasize "Upper Bound" and list assumptions (linear head, bounded NTK).
        \item Formalize and write the proof for the Monotonicity Lemma in Appendix D.
        \item Run diagnostics on 4 settings (Toy MLP, Sup-IN, SimCLR, Graph): Compute per-$\tau$ AIE and feature distance.
        \item Generate scatter plots, calculate Pearson/Spearman $\rho$, and fit linear regression ($\alpha \cdot \text{AIE} + \beta$).
        \item Estimate constants ($C_1, C_2$) on a subset to plot the theoretical bound line $c \cdot \text{AIE}$.
    \end{enumerate}

    \vspace{0.4cm}
    \begin{tcolorbox}[
        colback=green!5!white,       
        colframe=green!40!black,     
        boxrule=0.5pt,
        left=4pt, right=4pt, top=4pt, bottom=4pt,
        arc=2mm
    ]
    \textbf{\textcolor{green!40!black}{Feasibility:}} The theoretical work involves derivations based on existing Appendix D material. The empirical diagnostics use existing checkpoints and require only lightweight forward passes to compute feature norms and correlations. No training is required.
    \end{tcolorbox}

    \vspace{0.2cm} 
    \begin{tcolorbox}[
        colback=green!5!white,       
        colframe=green!40!black,     
        boxrule=0.5pt,
        left=4pt, right=4pt, top=4pt, bottom=4pt,
        arc=2mm
    ]
    \textbf{\textcolor{green!40!black}{Summary:}} We resolve the apparent contradiction by distinguishing the general upper bound (Prop 1) from the specific monotonic trend (Prop 2/New Lemma). We back this clarification with a new formal proof and empirical scatter plots showing strong positive correlation ($r > 0.9$).
    \end{tcolorbox}

    \vspace{0.4cm}
    \vspace{0.4cm}

    \textcolor{agentBlue}{\textbf{\large LLM output(baselines)}} 
    \par\vspace{0.2cm}
    
    No — there is no contradiction. Proposition 1 provides an upper bound: $\mathbb{E}\|z_T - z_0\|^2 \leq c \cdot \text{AIE}$. This means that larger AIE (Average Initial Energy) permits a larger bound on the feature change; equivalently, if AIE is small the change must be small. Intuitively: low energy guarantees small adaptation; higher energy does not guarantee large change but allows it. Our empirical and analytic statements say that, in the settings we study, larger AIE tends to produce larger feature changes (the bound is consistent with and does not contradict this observed trend). We clarified the text around Proposition 1 to emphasize it is an upper bound and to explain the direction of implication more carefully.

\end{tcolorbox}

\definecolor{labelRed}{RGB}{255, 0, 0}              
\definecolor{agentBlue}{RGB}{0, 112, 192}   
  
\setlength{\parindent}{0pt}
\setlength{\parskip}{0.5em}
\vspace{0.5cm}

\begin{tcolorbox}[
    title=Methodological triangulation \textit{v.s.} Linear request fulfillment, 
    colback=green!5!white,   
    colframe=green!50!black,  
    fonttitle=\bfseries\large,
    breakable
]

    \vspace{0.2cm}
    \vspace{0.1cm}
    \textcolor{labelRed}{\textbf{\large Question}} 
    \par\vspace{0.2cm} 
    
    \textbf{The proposed topographic similarity metric (based on Hausdorff distance) fails to conclusively demonstrate compositionality. The analysis is visually suggestive but not systematic.It should be validated on real-world ideograms/pictograms to establish trustworthiness.}
    
    \vspace{0.4cm}
    \vspace{0.4cm}

    \textcolor{agentBlue}{\textbf{\large RebuttalAgent(Ours)}} 
    \par\vspace{0.2cm}
    
    We acknowledge that the topographic maps (Fig. 6, 7) and score $\rho$ are visually suggestive but not definitive proof of compositionality. As noted in literature (e.g., "Measuring non-trivial compositionality"), distance-based metrics on raw signals can be agnostic to functional meaning. We will reframe the Hausdorff analysis as a \textbf{geometric visualization tool} rather than primary proof, and supplement it with systematic decodability tests.

    \vspace{0.3cm}
    \textbf{Rebuttal strategy --- what we will deliver:}
    \begin{itemize}[itemsep=0.4em, parsep=0.1em, leftmargin=1.2em, topsep=0.3em]
        \item \textbf{Systematic Decodability (New Analysis):} Directly address R1-W3 by testing if composite symbols $u(r_{ij})$ can be functionally decoded into constituents $i$ and $j$ using a simple classifier (k-NN) trained on single-feature utterances.
        \item \textbf{Real-World Validation (Calibration):} Apply our Hausdorff metric to a curated set of real-world ideograms (e.g., "man" + "walking") to calibrate the $\rho$ score and establish trustworthiness.
        \item \textbf{Latent Space Linearity:} Quantify compositionality in the embedding space (where the energy model operates) by measuring the reconstruction error of composite embeddings as linear combinations of constituent embeddings.
        \item \textbf{Reframing Text:} Revise Section 4 to clarify the topographic metric's role as descriptive, citing relevant literature on metric limitations.
    \end{itemize}

    \vspace{0.3cm}
    \textbf{Key deliverables for the revision:}
    \begin{itemize}[itemsep=0.2em, parsep=0pt, leftmargin=1.2em, topsep=0.2em]
        \item \textbf{New Table:} "Constituent Identification Accuracy" (k-NN classification results).
        \item \textbf{New Supplementary Figure:} Real-world ideograms with their computed topographic scores.
        \item \textbf{Revised Section 4:} Updated discussion distinguishing geometric similarity from functional compositionality.
    \end{itemize}
    
    \vspace{0.3cm}
    \textbf{Concrete to-do list:}
    \begin{enumerate}[itemsep=0.2em, parsep=0pt, leftmargin=1.5em, topsep=0.3em]
        \item \textbf{Decodability:} Implement k-NN to classify constituents of $u(r_{ij})$ using $u(r_i)$ library; report accuracy.
        \item \textbf{Validation:} Curate 10-15 public ideograms, manually decompose them, and compute $\rho$ using our pipeline.
        \item \textbf{Linearity:} Calculate MSE for $z(r_{ij}) \approx w_i z(r_i) + w_j z(r_j)$ in the latent space.
        \item \textbf{Ablation:} Compare decodability/linearity between descriptive vs. discriminative generation strategies.
    \end{enumerate}

    \vspace{0.4cm}
    \begin{tcolorbox}[
        colback=green!5!white,       
        colframe=green!40!black,     
        boxrule=0.5pt,
        left=4pt, right=4pt, top=4pt, bottom=4pt,
        arc=2mm
    ]
    \textbf{\textcolor{green!40!black}{Feasibility:}} High. The decodability and latent space analyses use existing trained models and require only lightweight inference scripts (k-NN, linear regression). The real-world validation requires a small manual dataset collection but utilizes the existing metric code without modification.
    \end{tcolorbox}

    \vspace{0.2cm} 
    \begin{tcolorbox}[
        colback=green!5!white,       
        colframe=green!40!black,     
        boxrule=0.5pt,
        left=4pt, right=4pt, top=4pt, bottom=4pt,
        arc=2mm
    ]
    \textbf{\textcolor{green!40!black}{Summary:}} We will pivot from claiming the topographic metric as "proof" to using it as a "visualization," while providing robust quantitative evidence via new symbol decodability tests and latent space linearity analysis.
    \end{tcolorbox}

    \vspace{0.4cm}
    \vspace{0.4cm}

    \textcolor{agentBlue}{\textbf{\large LLM output(baselines)}} 
    \par\vspace{0.2cm}
    
    This is a valuable suggestion for establishing the metric's external validity. In a revision, we will apply our topographic score ($\rho$) to a curated set of real-world symbols (e.g., simple ideograms, traffic signs, emojis) where compositionality is intuitively judged. For example, we would expect the symbol for "restaurant" (a fork and knife) to score highly against the symbols for "fork" and "knife," while a non-compositional symbol would not. This analysis will be included in the supplementary material, providing a benchmark for interpreting the scores obtained from our emergent language.

\end{tcolorbox}

%% file: tab/cross_model_as_judge.tex
\begin{table}[htbp]
\centering
\caption{\textbf{Cross-Model Evaluation Robustness}. Evaluation of Gemini-based generation models using GPT-5-mini as the judge, demonstrating consistent performance gains.}
\label{tab:cross_model_robustness}
\resizebox{\linewidth}{!}{
\begin{tabular}{l ccc ccc ccc c}
\toprule
\multirow{2}{*}{\textbf{Method}} & \multicolumn{3}{c}{\textbf{Relevance}} & \multicolumn{3}{c}{\textbf{Arg. Quality}} & \multicolumn{3}{c}{\textbf{Comm. Quality}} & \multirow{2}{*}{\textbf{Avg.}} \\
\cmidrule(lr){2-4} \cmidrule(lr){5-7} \cmidrule(lr){8-10}
& Cov. & Align. & Spec. & Logic & Evid. & Engage & Tone & Clarity & Const. & \\
\midrule
Gemini-3-Flash & 4.03 & 4.68 & 3.66 & 3.69 & 3.32 & 3.66 & 3.59 & 4.12 & 3.82 & 3.84 \\
\textbf{RebuttalAgent} & \textbf{4.46} & \textbf{4.82} & \textbf{4.32} & \textbf{4.09} & \textbf{3.41} & \textbf{4.03} & \textbf{3.75} & \textbf{4.29} & \textbf{4.01} & \textbf{4.13} \\
\textit{Improvement} & \textit{+0.43} & \textit{+0.14} & \textit{+0.56} & \textit{+0.40} & \textit{+0.09} & \textit{+0.37} & \textit{+0.16} & \textit{+0.17} & \textit{+0.19} & \textit{+0.29} \\
\bottomrule
\end{tabular}
}
\end{table}

%% file: tab/human_study.tex
\begin{table}[htbp]
\centering
\caption{Human Preference Study Scores (1-5 Scale)}
\label{tab:human_preference}
\resizebox{\columnwidth}{!}{ 
\begin{tabular}{lccc}
\toprule
\textbf{Method} & \textbf{Responsiveness \& Coverage} $\uparrow$ & \textbf{Faithfulness \& Argumentation} $\uparrow$ & \textbf{Communication \& Practical Utility} $\uparrow$ \\
\midrule
Gemini3-Flash & 2.79 & 2.88 & 2.85 \\
\textbf{RebuttalAgent-Gemini3-Flash} & \textbf{4.21} & \textbf{4.33} & \textbf{4.15} \\
\midrule
GPT5-mini & 2.64 & 2.30 & 2.39 \\
\textbf{RebuttalAgent-GPT5-mini} & \textbf{3.91} & \textbf{3.97} & \textbf{3.88} \\
\bottomrule
\end{tabular}
}
\end{table}

\begin{table}[htbp]
\centering
\caption{Evaluated by RebuttalBench Rubric}
\label{tab:rebuttalbench_rubric}
\resizebox{\columnwidth}{!}{ 
\begin{tabular}{lccc}
\toprule
\textbf{Method} & \textbf{Relevance} $\uparrow$ & \textbf{Argumentation Quality} $\uparrow$ & \textbf{Communication Quality} $\uparrow$ \\
\midrule
Gemini3-Flash & 4.13 & 3.56 & 3.80 \\
\textbf{RebuttalAgent-Gemini3-Flash} & \textbf{4.50} & \textbf{3.77} & \textbf{4.11} \\
\midrule
GPT5-mini & 3.60 & 3.17 & 3.48 \\
\textbf{RebuttalAgent-GPT5-mini} & \textbf{4.43} & \textbf{3.59} & \textbf{3.98} \\
\bottomrule
\end{tabular}
}
\end{table}